\documentclass{article}
\usepackage{graphicx}
\usepackage{caption}
\usepackage{hyperref}
\usepackage{listings}
\usepackage{xcolor}
\usepackage{amsmath}
\usepackage{algorithm}
\usepackage{algpseudocode}
\usepackage{amsthm}
\usepackage{tikz}
\usetikzlibrary{positioning}
\usetikzlibrary{arrows}
\usetikzlibrary{positioning,calc}
\usepackage{subcaption}
\usepackage{authblk}

\begin{document}

\title{\Huge A better method to enforce monotonic constraints in regression and classification trees}
%\author{Auguste, Charles \and Smirnov, Ivan \and Malory, Sean}
\author[1]{Auguste, Charles}
\author[2]{Smirnov, Ivan}
\author[2]{Malory, Sean}
\affil[1]{IMI - Departement Ingenierie Mathematique et Informatique, Ecole des Ponts ParisTech}
\affil[2]{Independent Researcher}

\maketitle

\begin{abstract}
In this report we present two new ways of enforcing monotone constraints in regression and classification trees.
One yields better results than the current LightGBM, and has a similar computation time.
The other one yields even better results, but is much slower than the current LightGBM.
We also propose a heuristic that takes into account that greedily splitting a tree by choosing a monotone split with respect to its
immediate gain is far from optimal.
Then, we compare the results with the current implementation of the constraints in the LightGBM library,
using the well known Adult public dataset.
    Throughout the report, we mostly focus on the implementation of our methods that we made for the LightGBM library,
even though they are general and could be implemented in any regression or classification tree. \\

    The best method we propose (a smarter way to split the tree coupled to a penalization of monotone splits)
    consistently beats the current implementation of LightGBM.
    With small or average trees, the loss reduction can be as high as 1\% in the early stages of training and decreases to around 0.1\% at the loss peak for the Adult dataset.
    The results would be even better with larger trees.
    In our experiments, we didn't do a lot of tuning of the regularization parameters, and we wouldn't be surprised to see that increasing the performance of our methods on test sets.
\end{abstract}

    \section{Results} \label{res}

    \subsection{Summary of the important results}

\begin{figure}[H]
\makebox[\textwidth][c]{
\includegraphics[width=.85\textwidth]{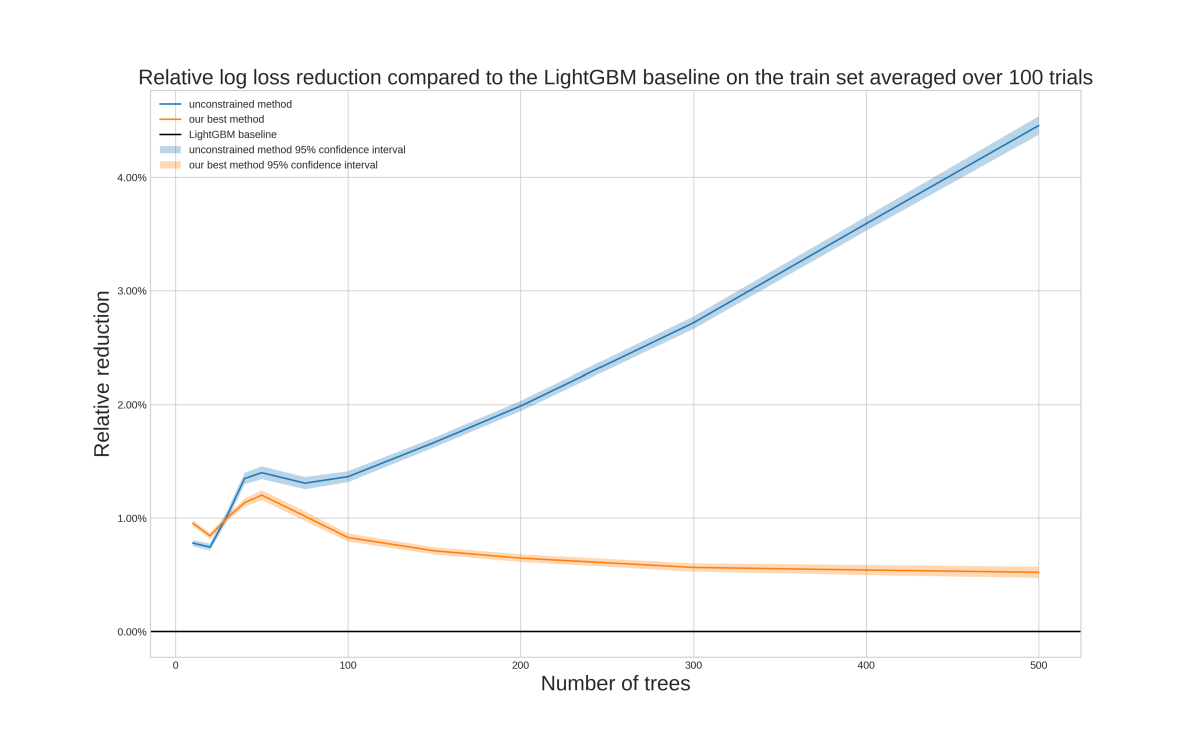}\hfill
\includegraphics[width=.85\textwidth]{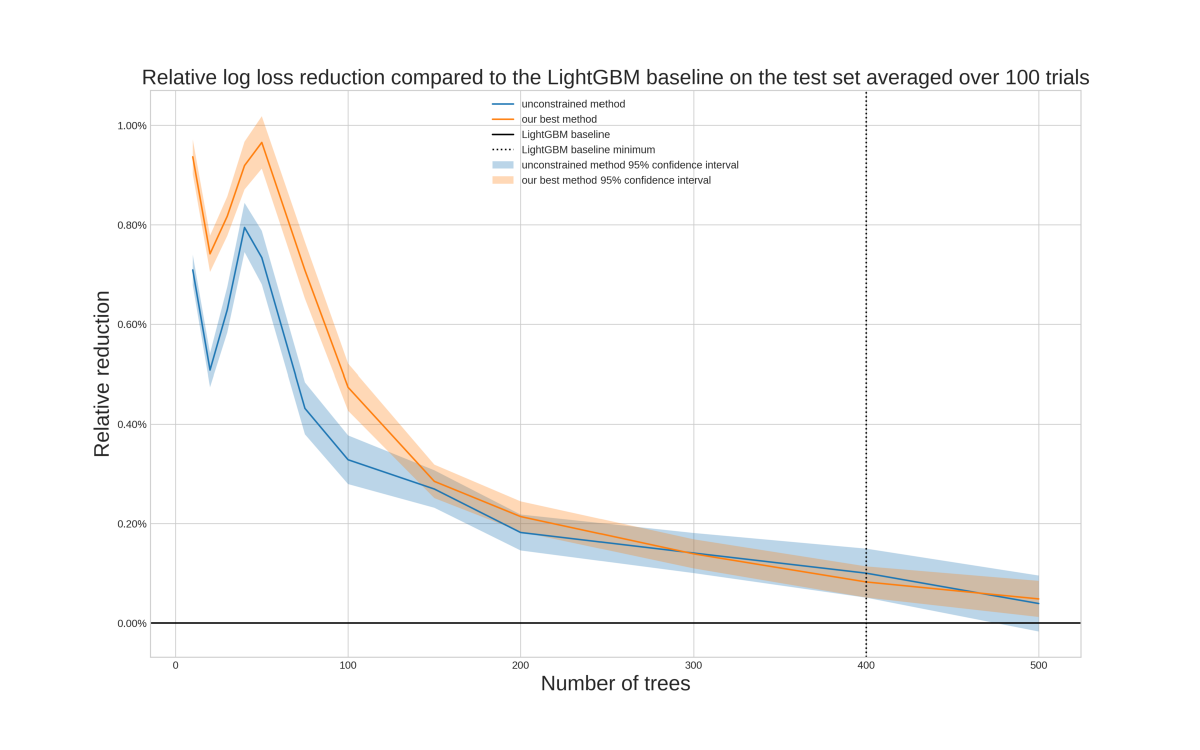}\hfill
}
\makebox[\textwidth][c]{
\includegraphics[width=.85\textwidth]{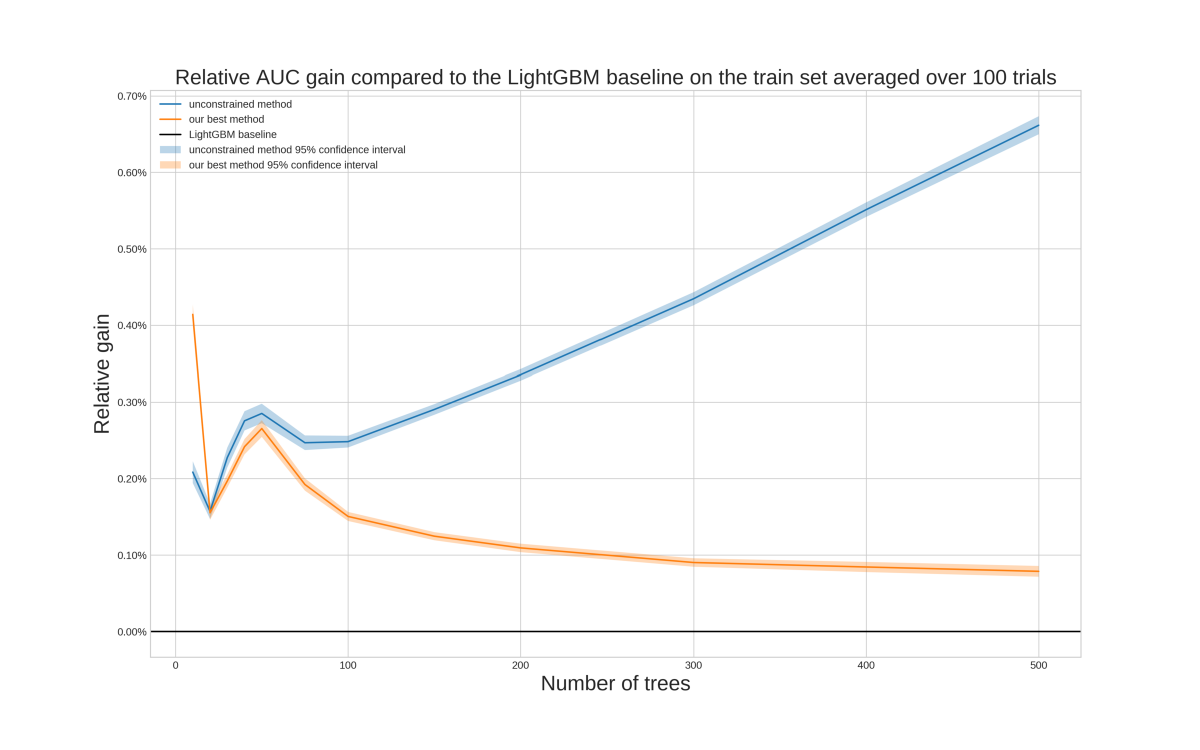}\hfill
\includegraphics[width=.85\textwidth]{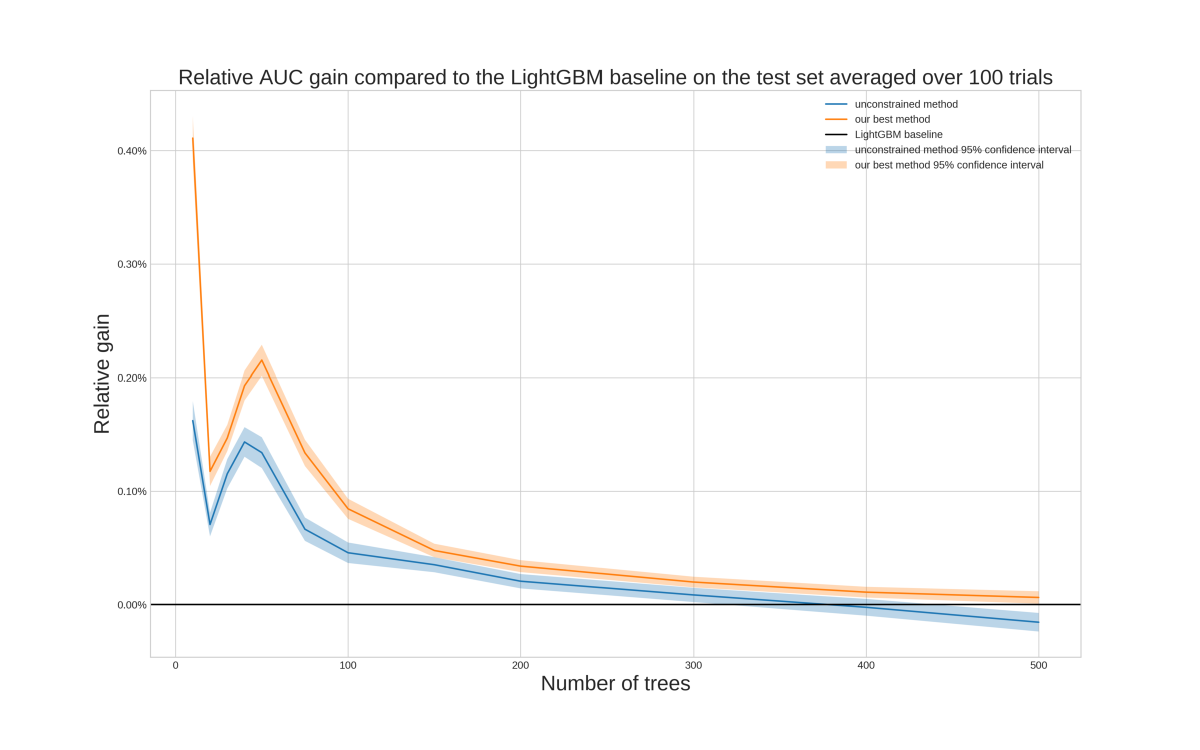}\hfill
}
\captionsetup{justification=centering,margin=2cm}
\caption{Loss and AUC vs. number of iterations relative to the constrained LightGBM baseline, for our best method and the unconstrained method, on the train and test sets}
\label{imp}
\end{figure}

On figure \ref{imp} are represented the most important results of the report.
Overall, our best method generates a consistent and significant loss reduction, which also helps us improve common metrics.
Therefore we think our method should become the standard for enforcing monotone constraints,
and replace the current LightGBM method.

    \subsection{Dataset and preprocessing}

The Adult dataset is a well known public dataset extracted by Barry Becker from the 1994 Census database
\footnote{\href{https://archive.ics.uci.edu/ml/datasets/adult}{Adult data set} Dua, D. and Graff, C. (2019). UCI Machine Learning Repository [http://archive.ics.uci.edu/ml]. Irvine, CA: University of California, School of Information and Computer Science.}.
The task of this dataset is classification.
We need to classify individuals depending on whether or not they earn more than 50,000\$ a year.
The dataset is fairly imbalanced as for around 75\% of entries, people earn less than 50,000\$ a year
(which doesn't mean that around 75\% of people earned less than 50,000\$ in 1994, because entries have to be weighted by the survey weights).
If a person earns more than 50,000\$ a year, its label will be 1.
Otherwise, it will be 0.

\paragraph*{Description of the data}
The description of the data is extremely important as we need to know which variable are locally monotone.
Table \ref{adultDescription} summarizes for each variable its type (continuous or discrete) and if the labels should be locally monotone with respect to it.

\begin{table}[h]
\begin{center}
\begin{tabular}{ |c|c|c| }
 \hline
 Variable & Type & Labels' relationship to the variable \\
 \hline
 \texttt{age} & continuous & Monotonically increasing \\
 \texttt{workclass} & discrete & None \\
 \texttt{fnlwgt} & continuous & None \\
 \texttt{education} & discrete & None \\
 \texttt{education\_num} & continuous & Monotonically increasing \\
 \texttt{marital\_status} & discrete & None \\
 \texttt{occupation} & discrete & None \\
 \texttt{relationship} & discrete & None \\
 \texttt{race} & discrete & None \\
 \texttt{sex} & discrete & None \\
 \texttt{capital\_gain} & continuous & None \\
 \texttt{capital\_loss} & continuous & None \\
 \texttt{hours\_per\_week} & continuous & Monotonically increasing \\
 \texttt{native\_country} & discrete & None \\
 \hline
\end{tabular}
\end{center}
    \caption{Summary of the Adult dataset variables}
    \label{adultDescription}
\end{table}

\paragraph*{Treatment of the data}
The variables \texttt{education} and \texttt{education\_num} are bijective.
We chose to remove the \texttt{education} variable as the labels are monotonically increasing with respect to the other one.
Then, we chose to one-hot encode all the categorical variables. \\

In order to test the different algorithms, de decided to perform Monte-Carlo cross validation (or repeated random sub-sampling validation).
For each experiment, we perform $N$ trials where we randomly split the set into a training set and a testing set.
The results are then averaged across all the trials.
The ratios we are using are shown on table \ref{setSplits}.

\begin{table}[h]
\begin{center}
\begin{tabular}{ |c|c|c| }
 \hline
  Set & Ratio & Number of data points \\
 \hline
  Training set & 65\% & 31747 \\
  Testing set & 35\% & 17095 \\
 \hline
    Total & 100\% & 48842 \\
 \hline
\end{tabular}
\end{center}
    \caption{Split ratio of the data into training and testing}
    \label{setSplits}
\end{table}

\paragraph*{Parameters used}
The parameters we used for our experiments were set as stated in table \ref{params}.
The remaining parameters not mentioned in table \ref{params} were set to LightGBM's default values\footnote{\href{https://lightgbm.readthedocs.io/en/latest/Parameters.html}{LightGBM parameters}}.
Finally, we set the variables \texttt{age}, \texttt{education\_num} and \texttt{hours\_per\_weeks} to be monotonically increasing.
We decided to perform gradient boosting.
We picked the specific depth 5 to have trees that wouldn't overfit too much, otherwise it would make our new methods perform better
than what they actually do on the training set and worse on the test set (because they will be able to overfit better as they are less "constraining").

\begin{table}[h]
\begin{center}
\begin{tabular}{ |c|c|c| }
 \hline
  Parameter & Value \\
 \hline
  objective & \texttt{"binary"} \\
  boosting & \texttt{"gbdt"} \\
  num\_leaves &  32 \\
  depth & 5 \\
  min\_data\_in\_leaf & 100 \\
  seed & 42 \\
  bagging\_seed & 42 \\
  feature\_fraction\_seed & 42 \\
  drop\_seed & 42 \\
  data\_random\_seed & 42 \\
 \hline
\end{tabular}
\end{center}
    \caption{Parameters chosen}
    \label{params}
\end{table}

    \subsection{Loss and metrics vs. iterations}
The unconstrained method represents an upper bound of the achievable performance on the training set (we cannot learn better with constraints than without).
We use the current constrained LightGBM implementation as a baseline and plot our results according to it. \\

On figure \ref{logloss}, it can be seen that our new methods learn more efficiently than the LightGBM baseline and
consistently beat it on the training set for the loss and for all metrics.
On the testing set, we also mostly beat the baseline.
However, with many iterations we start to overfit, and therefore some metrics can get below the baseline, but this could be
offset by tuning the regularization parameters of the gradient boosting scheme. \\

On these figures, it seems like the slow method is not that useful as a combination of the fast method and penalization achieves almost the same results.
However, for different datasets, and different parameters, especially if we were trying to build bigger trees, then the slow method could achieve significantly better results.

\begin{figure}[htb]
\makebox[\textwidth][c]{
\includegraphics[width=.85\textwidth]{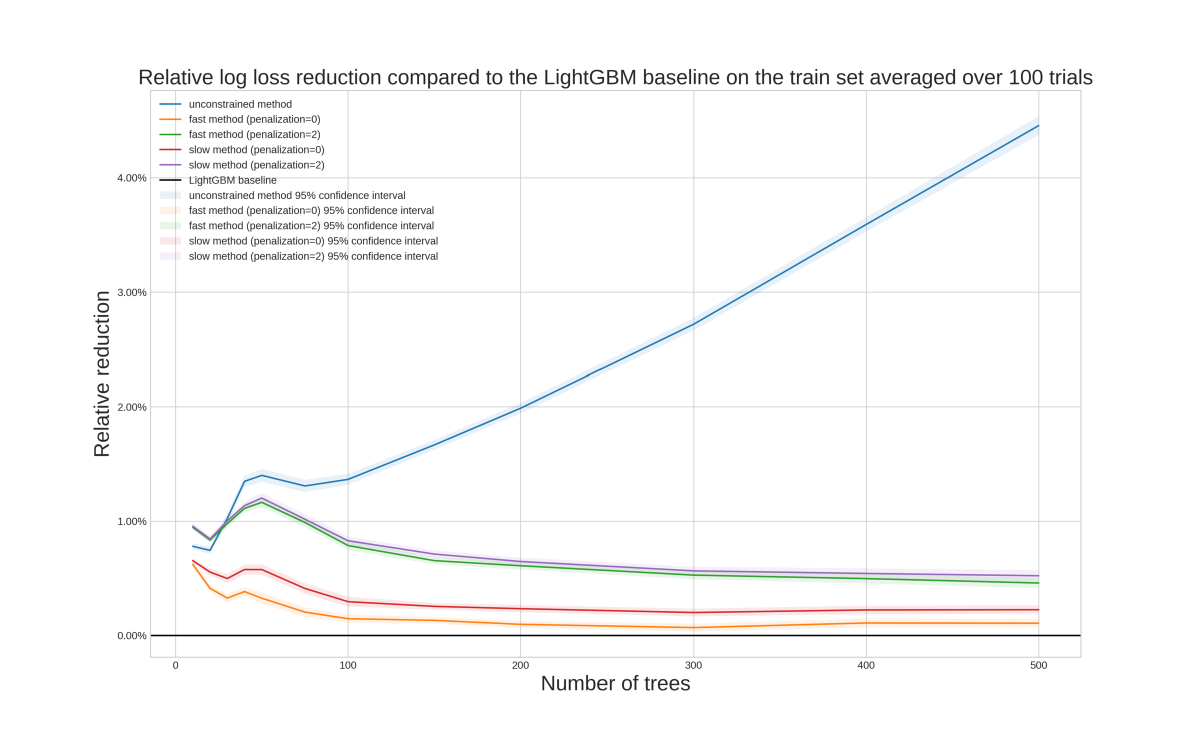}\hfill
\includegraphics[width=.85\textwidth]{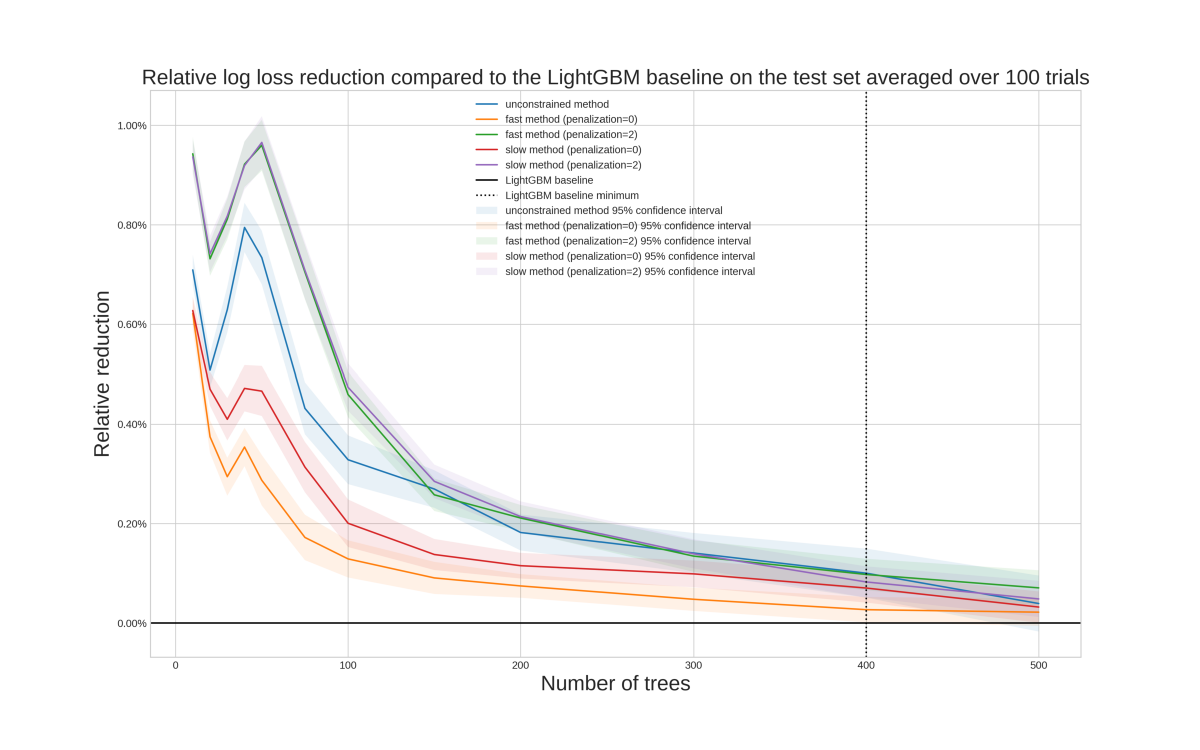}\hfill
}
\makebox[\textwidth][c]{
\includegraphics[width=.85\textwidth]{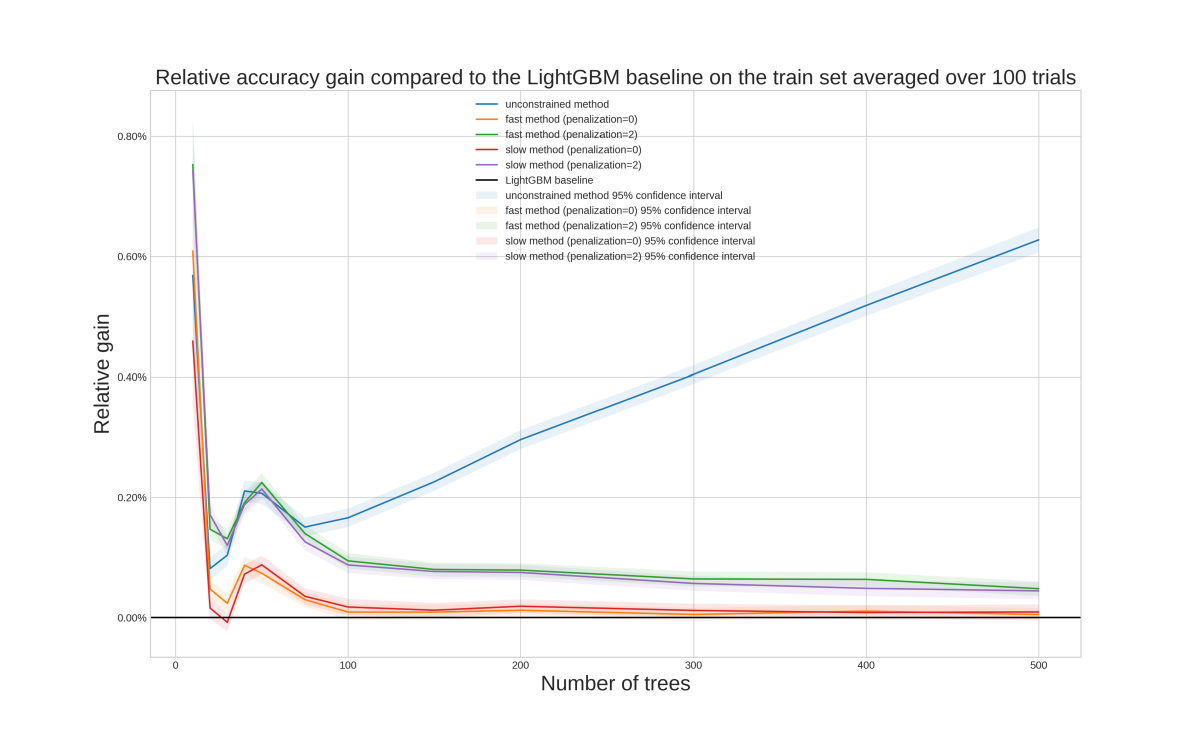}\hfill
\includegraphics[width=.85\textwidth]{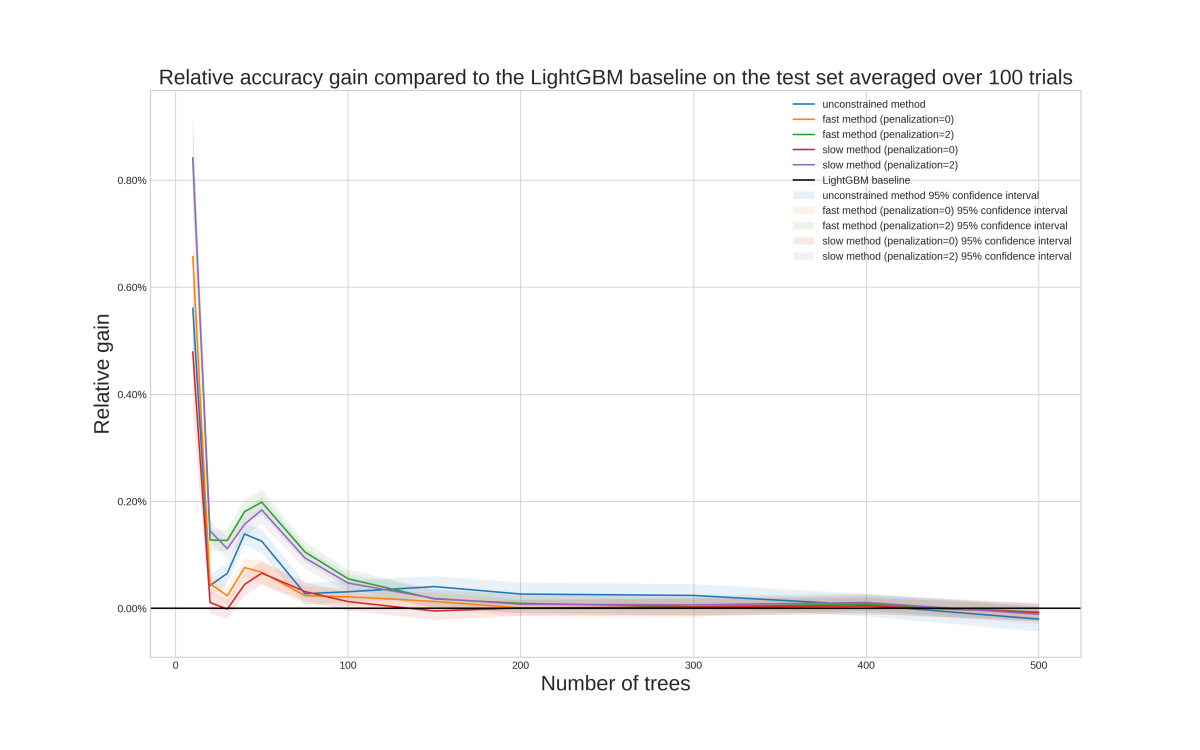}\hfill
}
\captionsetup{justification=centering,margin=2cm}
\caption{Loss and metrics vs. number of iterations relative to the constrained LightGBM baseline, for all methods, on the train and test sets}
\end{figure}
\begin{figure}[htb]\ContinuedFloat
\makebox[\textwidth][c]{
\includegraphics[width=.85\textwidth]{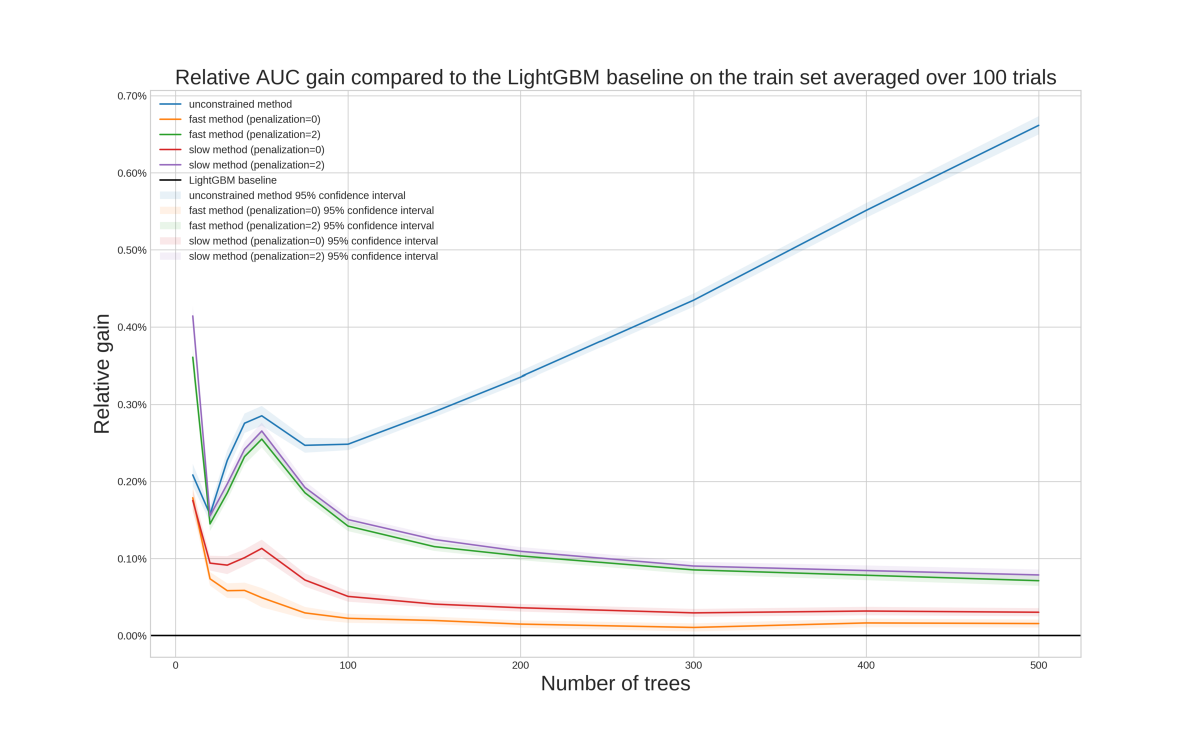}\hfill
\includegraphics[width=.85\textwidth]{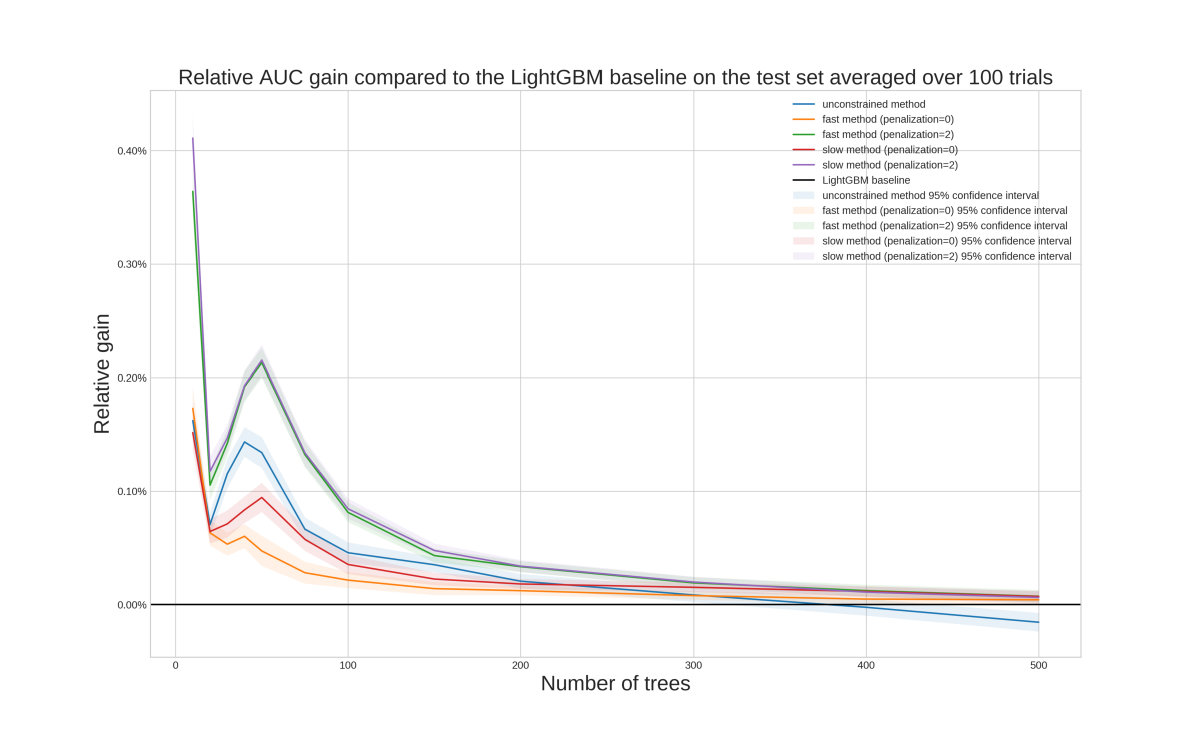}\hfill
}
\captionsetup{justification=centering,margin=2cm}
\caption{Loss and metrics vs. number of iterations relative to the constrained LightGBM baseline, for all methods, on the train and test sets}
\label{logloss}
\end{figure}

    \subsection{Stability of the loss for with respect to the penalization parameter}

\begin{figure}[htp]
\centering
\includegraphics[width=1.\textwidth]{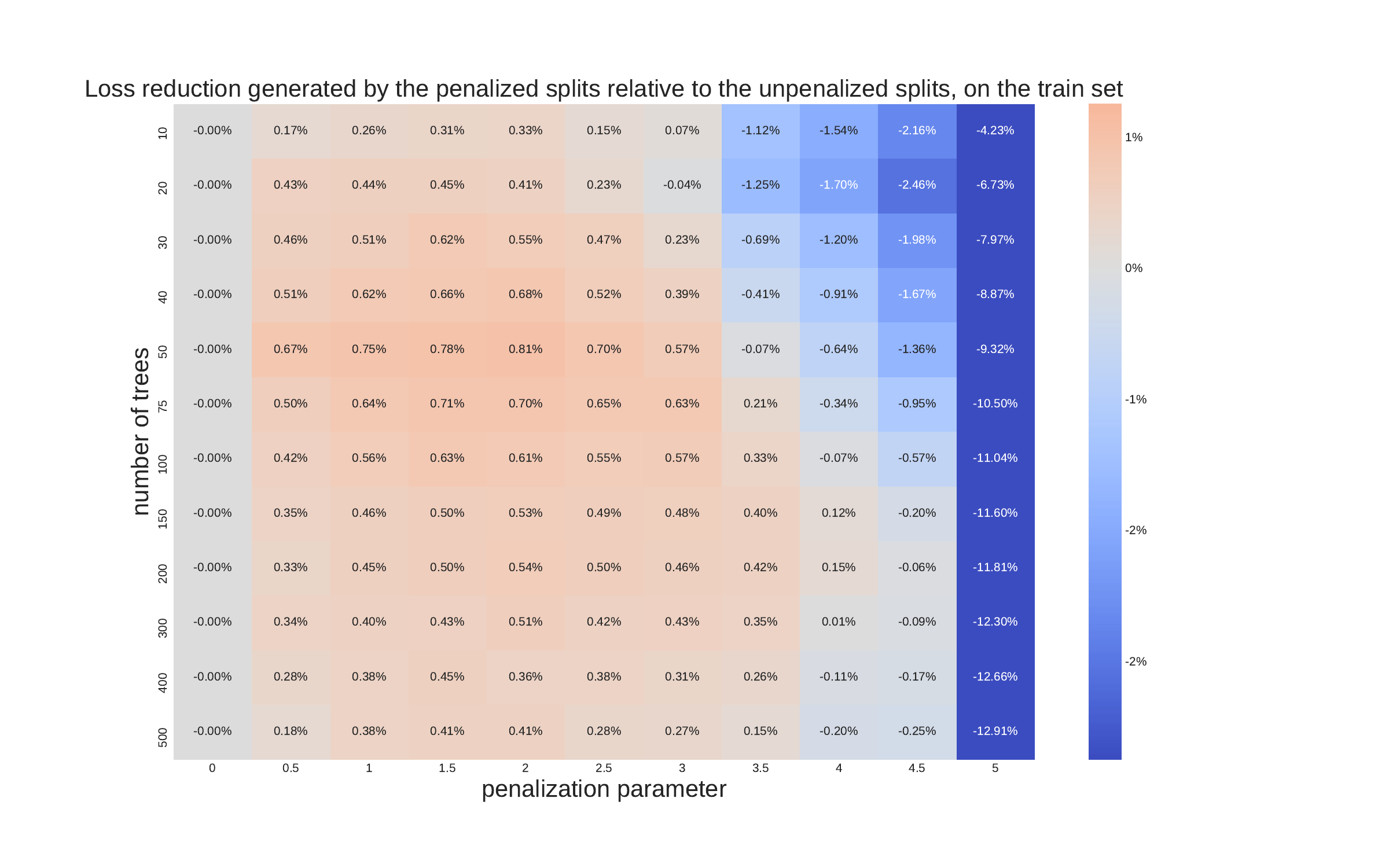}\hfill
\captionsetup{justification=centering,margin=2cm}
\caption{Relative loss reduction for penalized monotone splits, compared to unpenalized splits, for different penalization parameters, fot the fast method}
\label{iterations_gamma_fast}
\end{figure}

On figure \ref{iterations_gamma_fast}, it can be observed that our heuristic penalty is able to reduce the loss when it is set correctly.
Precise tuning is not required as the results seem pretty stable, and the improvement is consistent (even though it is offset by the boosting effect).
However, one must be careful to not set the penalization parameter too high, otherwise, it may prevent any monotone split in the whole tree.
Figure \ref{iterations_gamma_fast} only depicts the results for the fast method.
However, the slow method yields very similar results.

    \subsection{Computation time for each method}

\begin{figure}[htp]
\centering
\includegraphics[width=1.\textwidth]{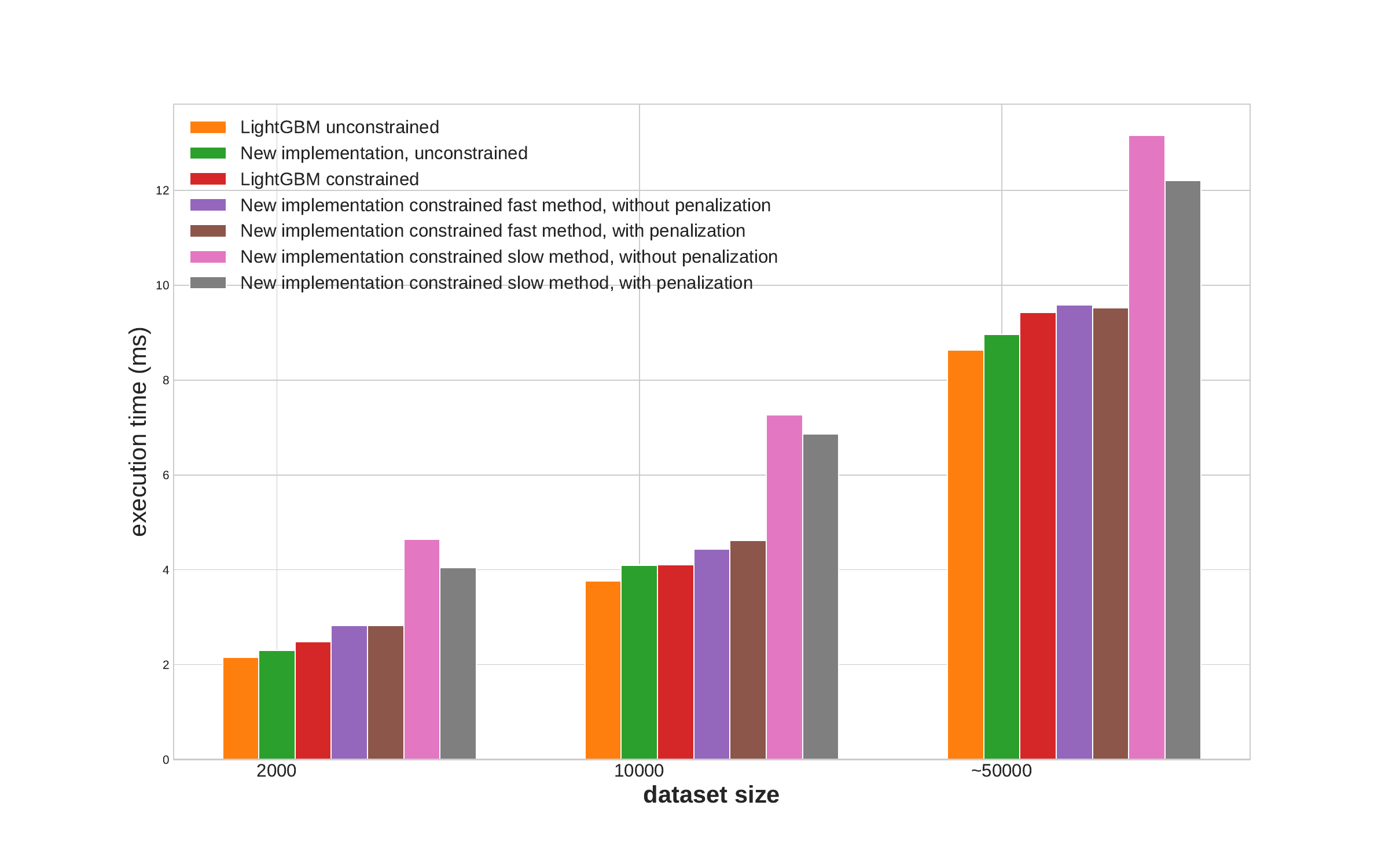}\hfill
\captionsetup{justification=centering,margin=2cm}
\caption{Mean execution time of the different methods for different dataset sizes averaged over 5000 runs}
\label{timings}
\end{figure}

On figure \ref{timings}, we report the computational cost for each method.
We measured it using the \texttt{timeit} module in Python
(so the results are not very precise, as the standard deviation can be as high as 1ms).
These statistics were created using either 2000, 10000, or close to 50000 random entries.
The execution times were measured over 1 boosting iteration, using a maximum depth of 10 and a maximum number of leaves of 40, and the trees were filled with the maximum number of leaves every time.
The results are averaged across 5000 runs.\\

Even though these results are a bit noisy, we can draw very useful insights from them.
Overall, whether we are using LightGBM or our new fast method,
the difference in computational costs is not very significant.
The slow method is, however, much slower than the current LightGBM.
Nonetheless, the difference mitigates when the dataset is bigger and building histogram is a more important task.

\subsection{Plots of the first trees generated by each method}

\begin{figure}[htp]
\centering
\includegraphics[width=1.\textwidth]{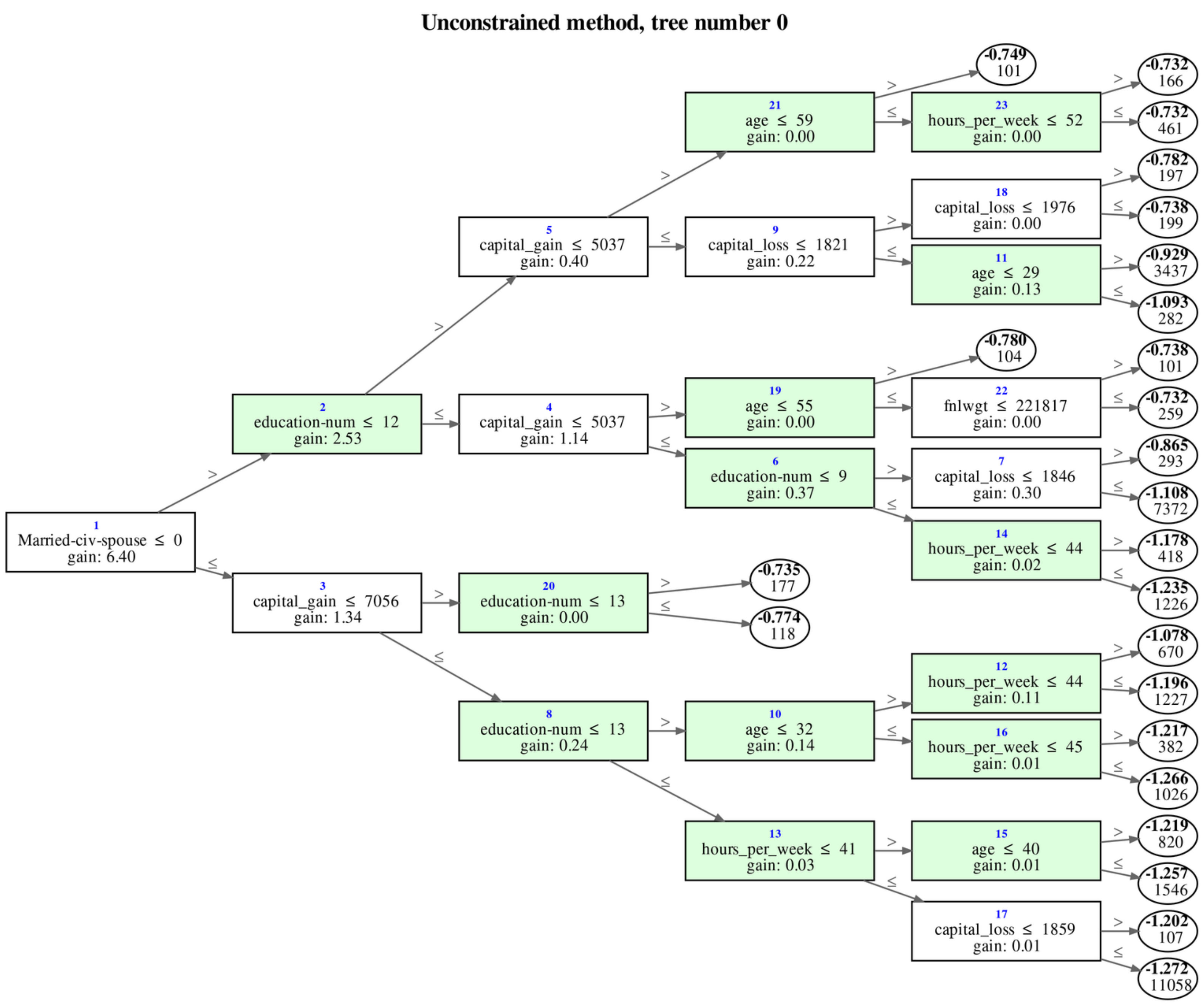}\hfill
\includegraphics[width=1.\textwidth]{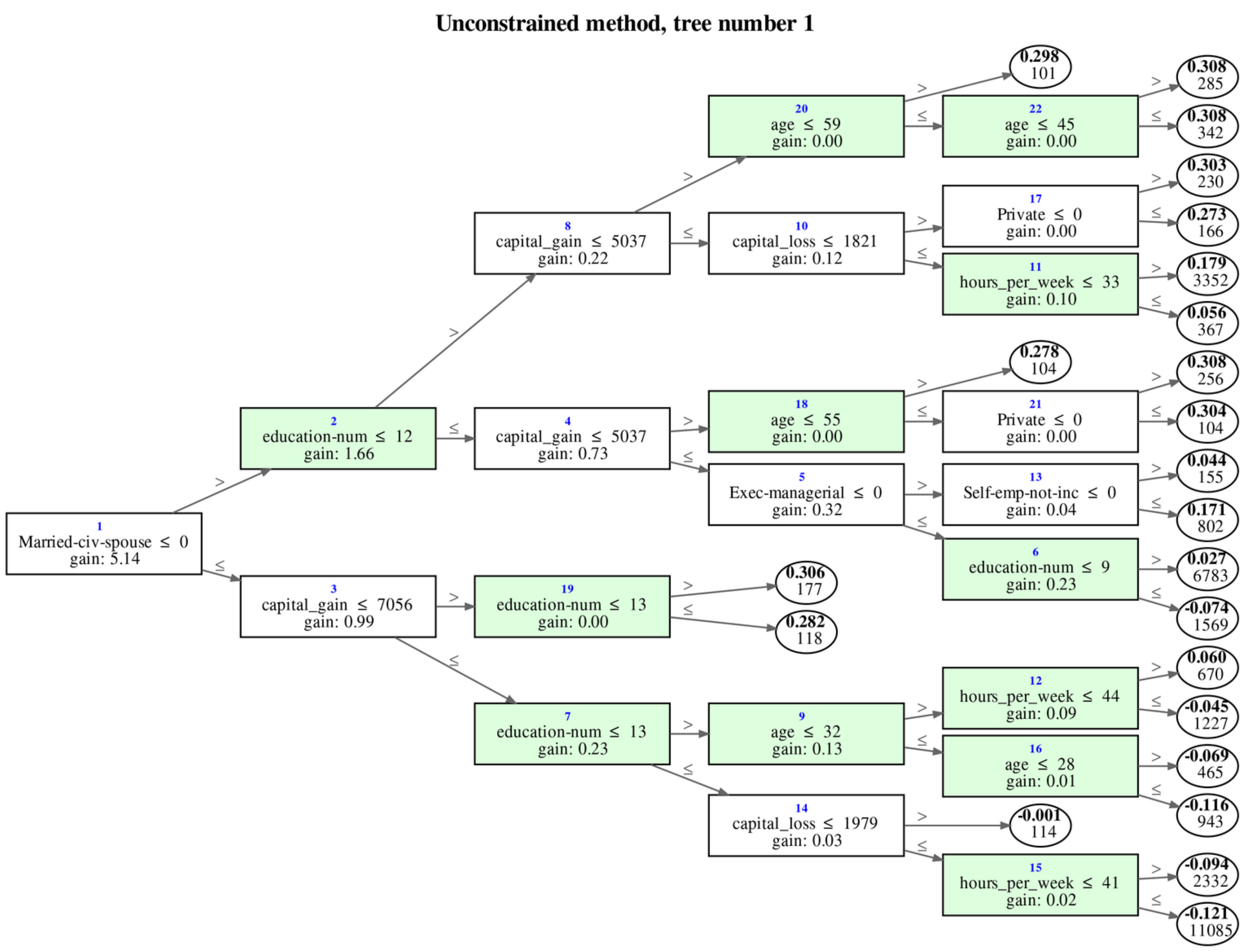}\hfill
\captionsetup{justification=centering,margin=2cm}
\caption{First two trees generated by the unconstrained method}
\label{tree_unconstrained}
\end{figure}

\begin{figure}[htp]
\centering
\includegraphics[width=1.\textwidth]{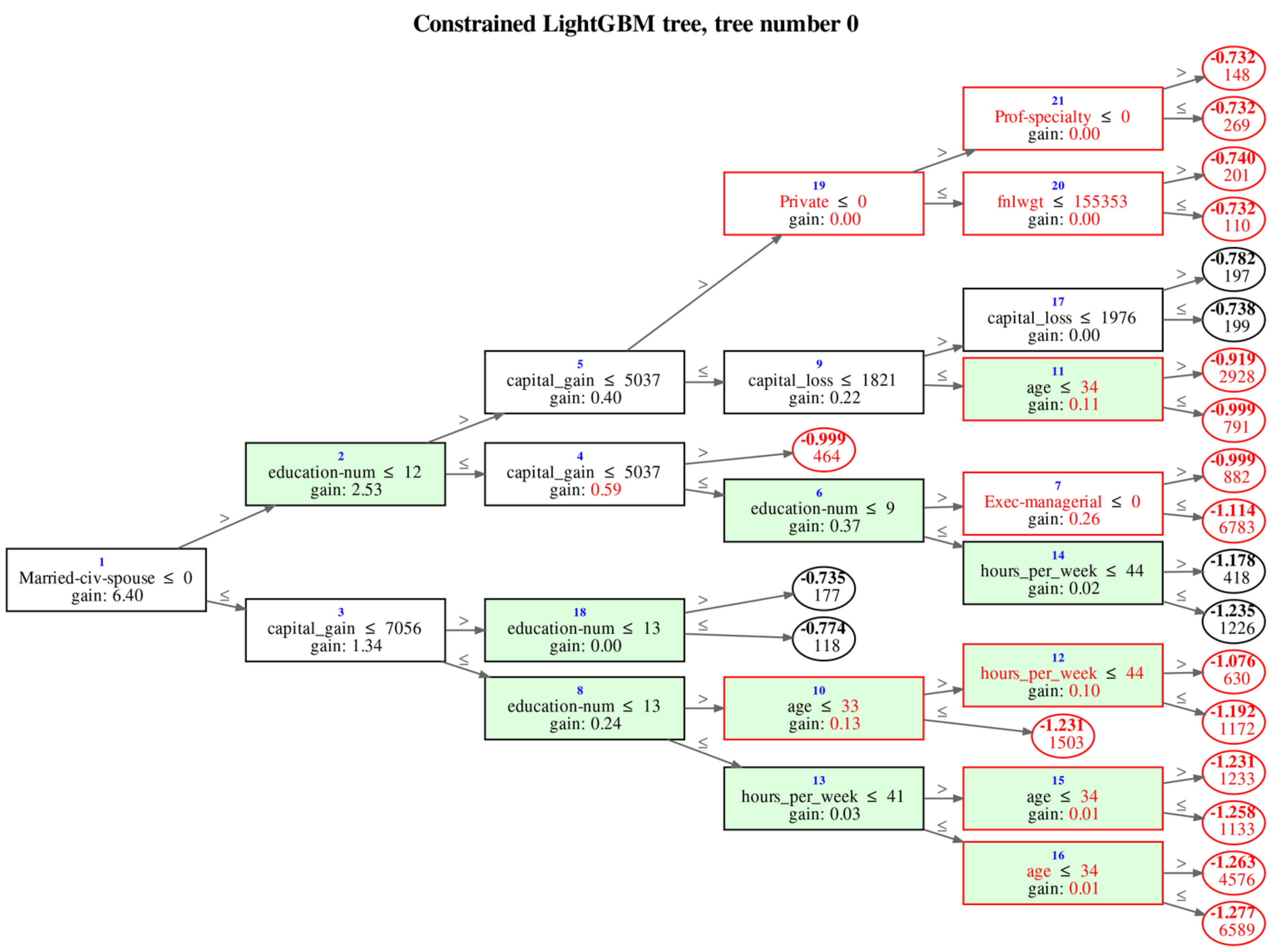}\hfill
\includegraphics[width=1.\textwidth]{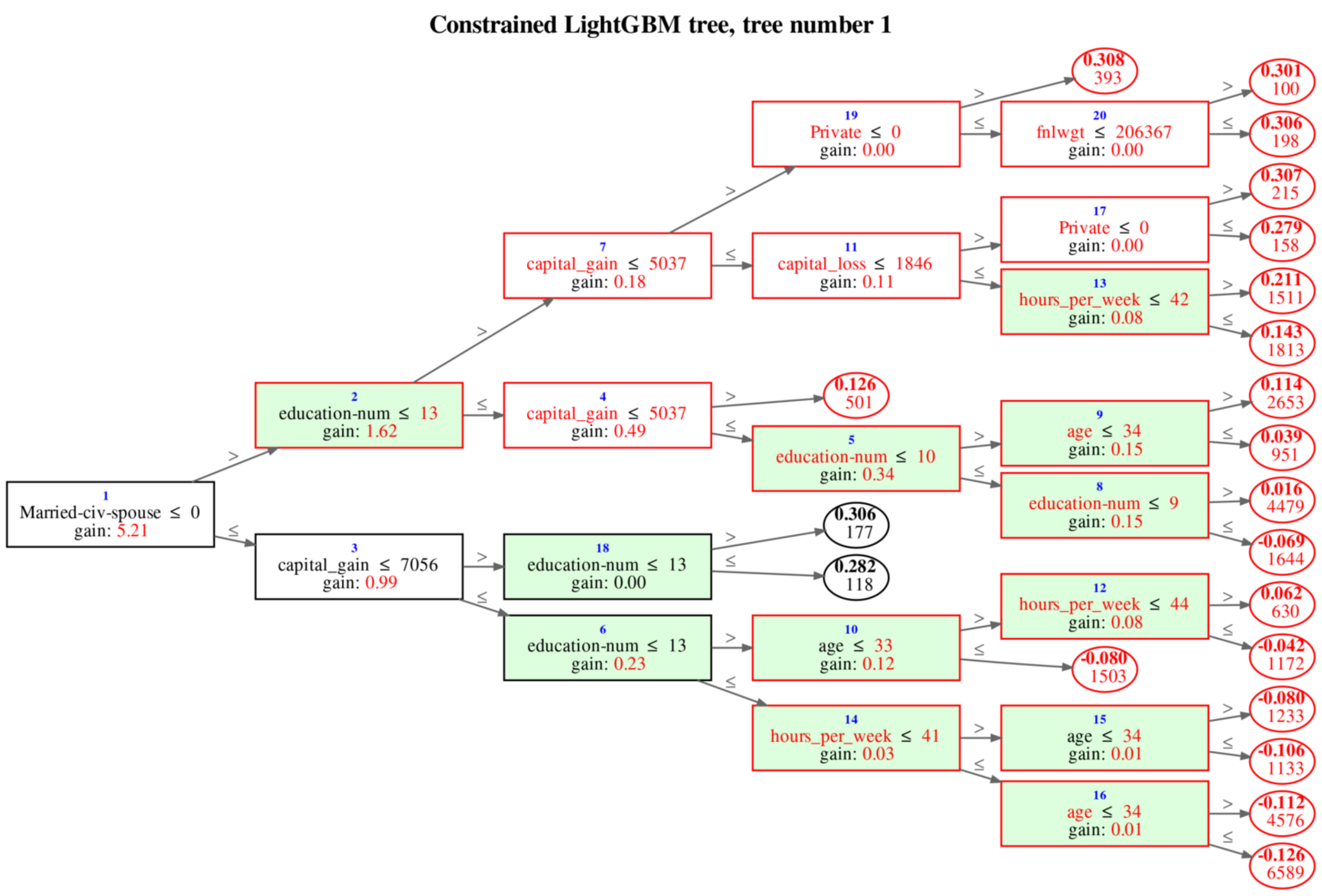}\hfill
\captionsetup{justification=centering,margin=2cm}
\caption{First two trees generated by the baseline LightGBM method}
\label{tree_lightgbm}
\end{figure}

\begin{figure}[htp]
\centering
\includegraphics[width=1.\textwidth]{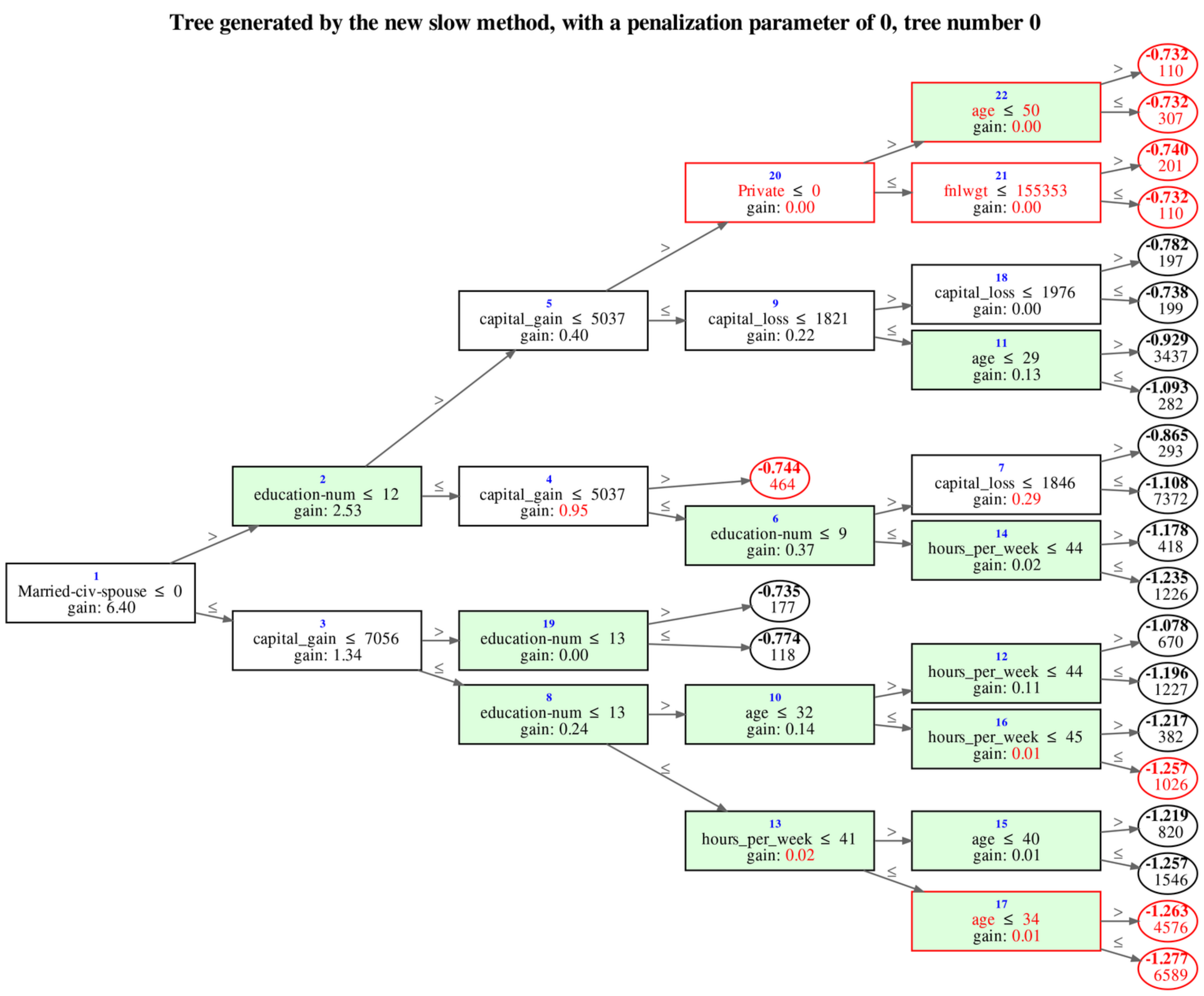}\hfill
\includegraphics[width=1.\textwidth]{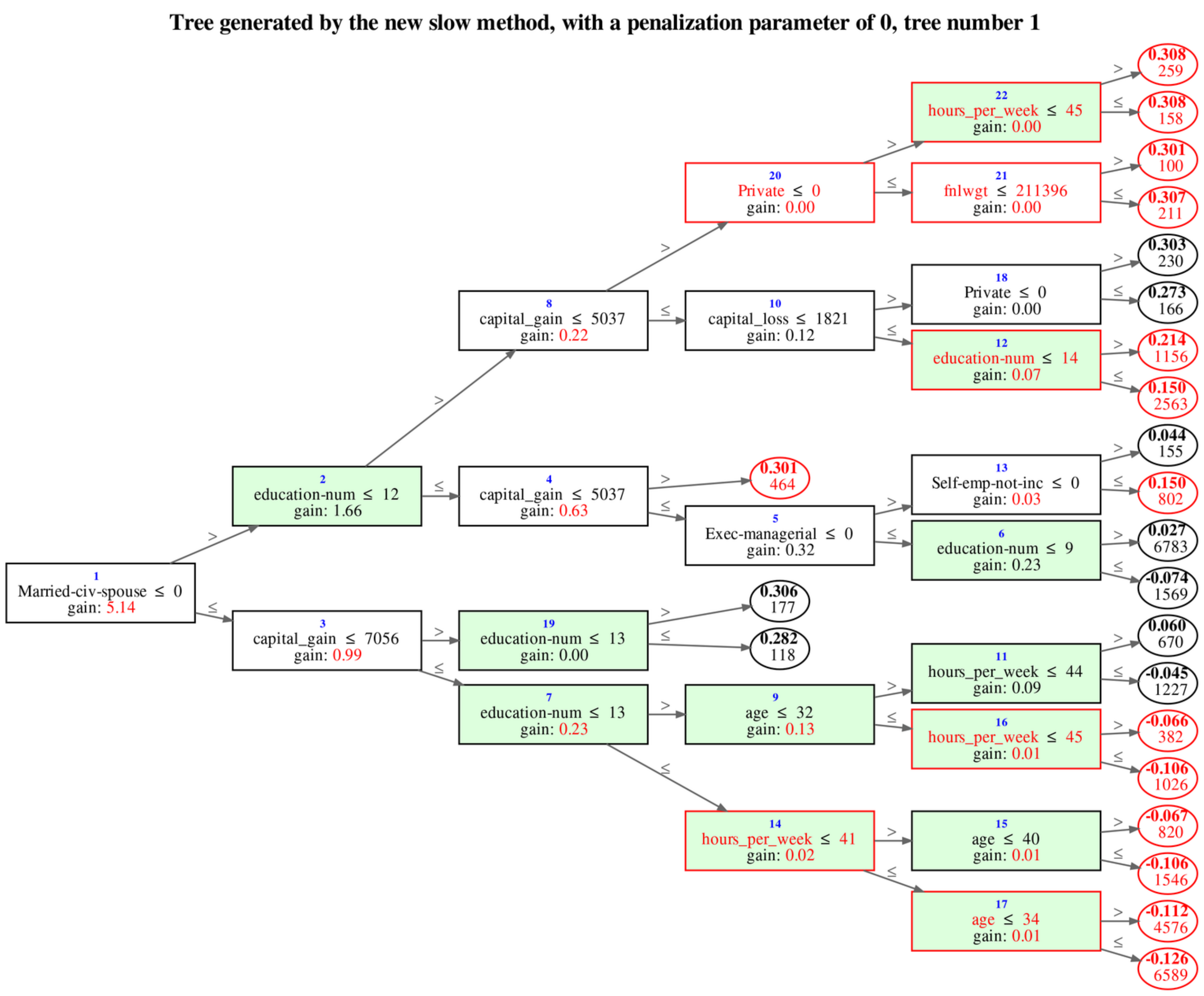}\hfill
\captionsetup{justification=centering,margin=2cm}
\caption{First two trees generated by our new slow method, with no penalization}
\label{tree_slow}
\end{figure}

\begin{figure}[htp]
\centering
\includegraphics[width=1.\textwidth]{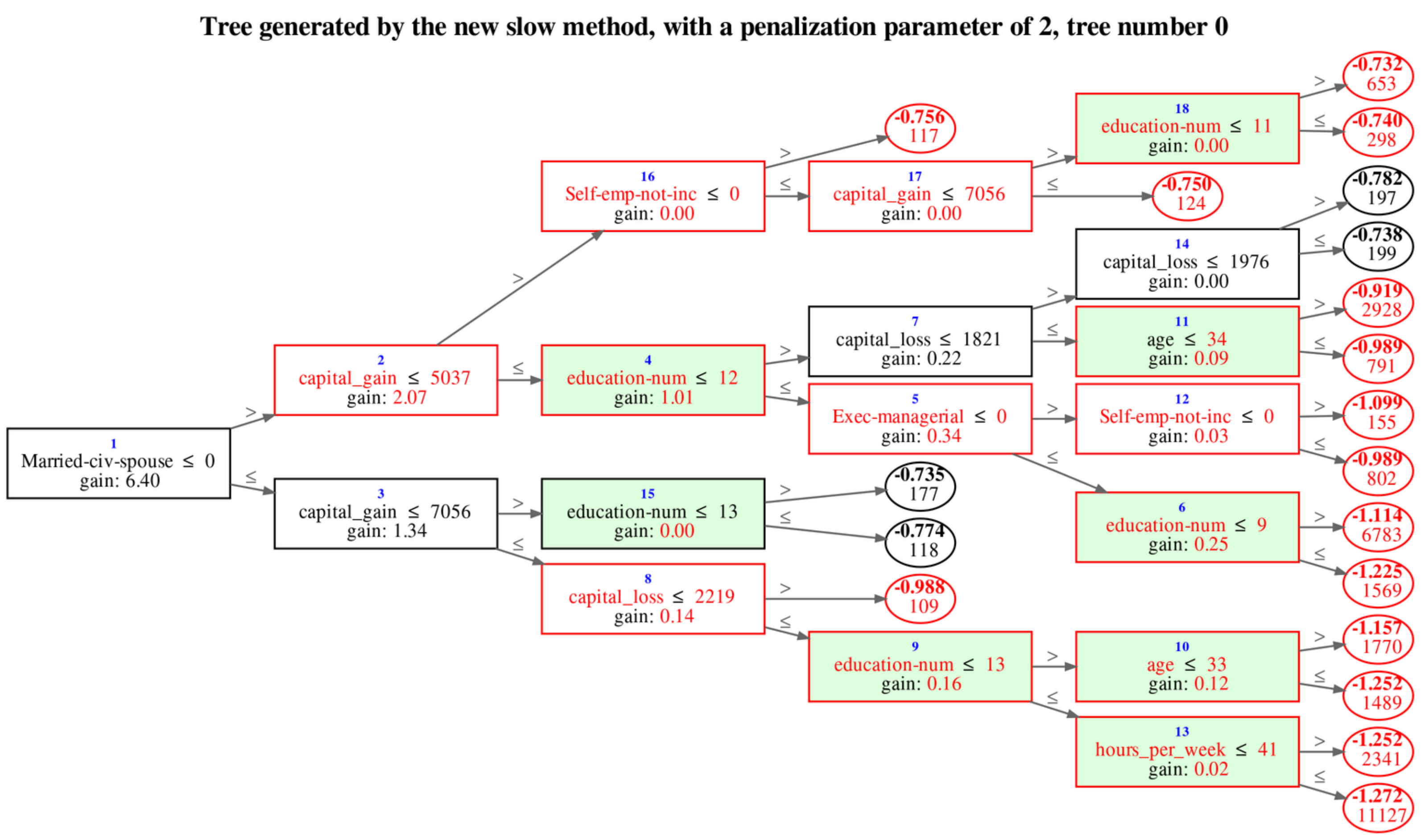}\hfill
\includegraphics[width=1.\textwidth]{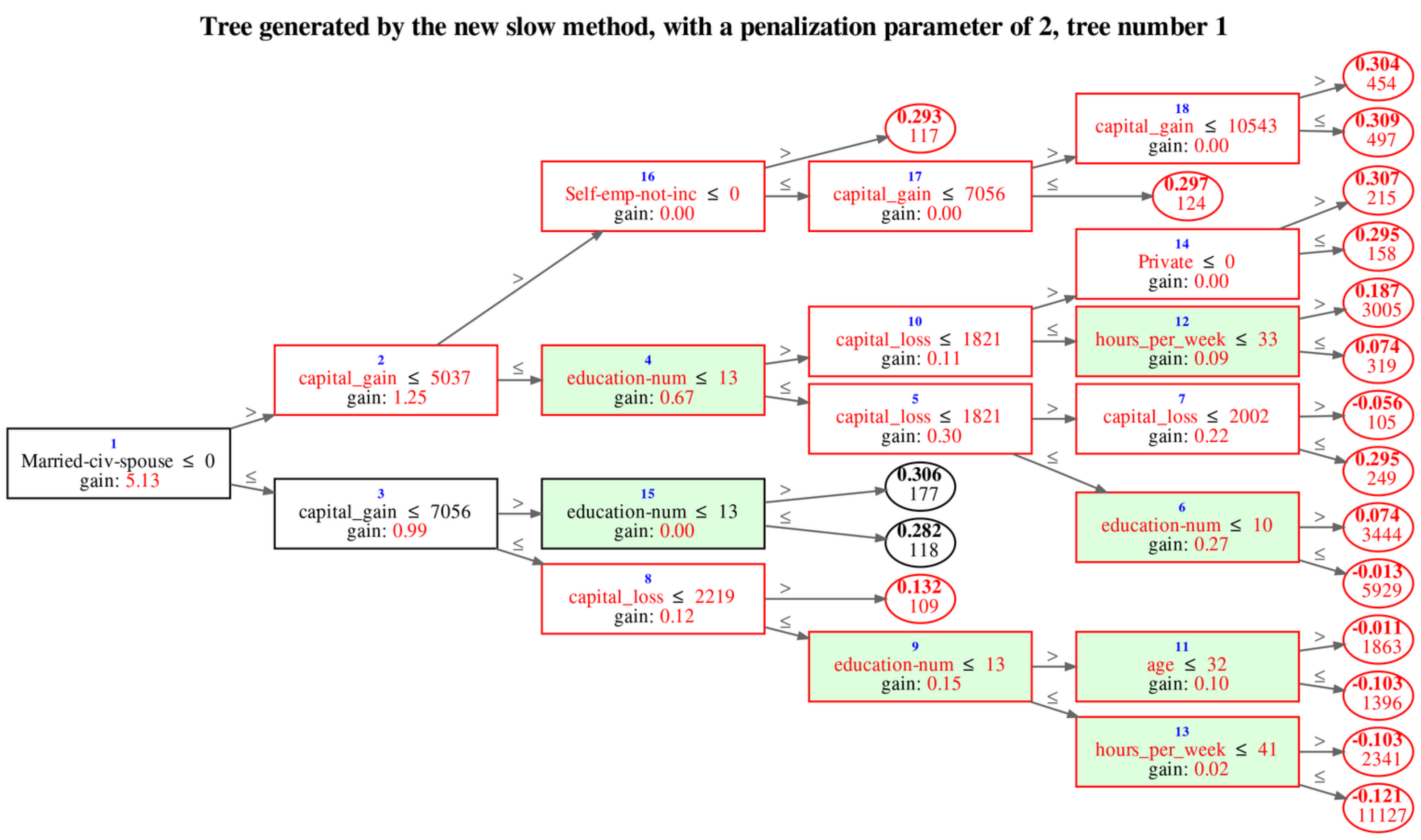}\hfill
\captionsetup{justification=centering,margin=2cm}
\caption{First two trees generated by our new slow method, with a penalization parameter of 2}
\label{tree_slow_penalize}
\end{figure}

\begin{figure}[htp]
\centering
\includegraphics[width=1.\textwidth]{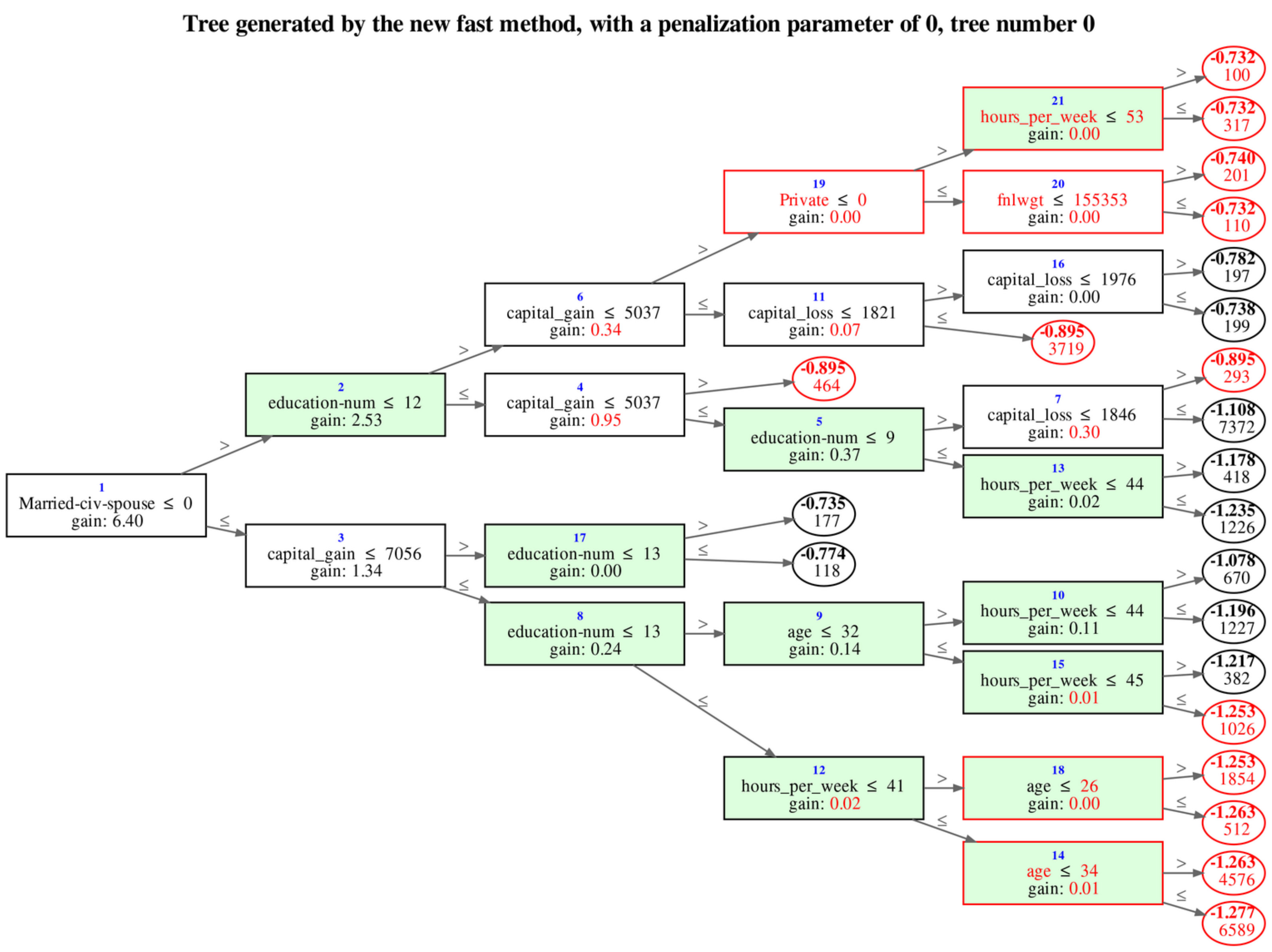}\hfill
\includegraphics[width=1.\textwidth]{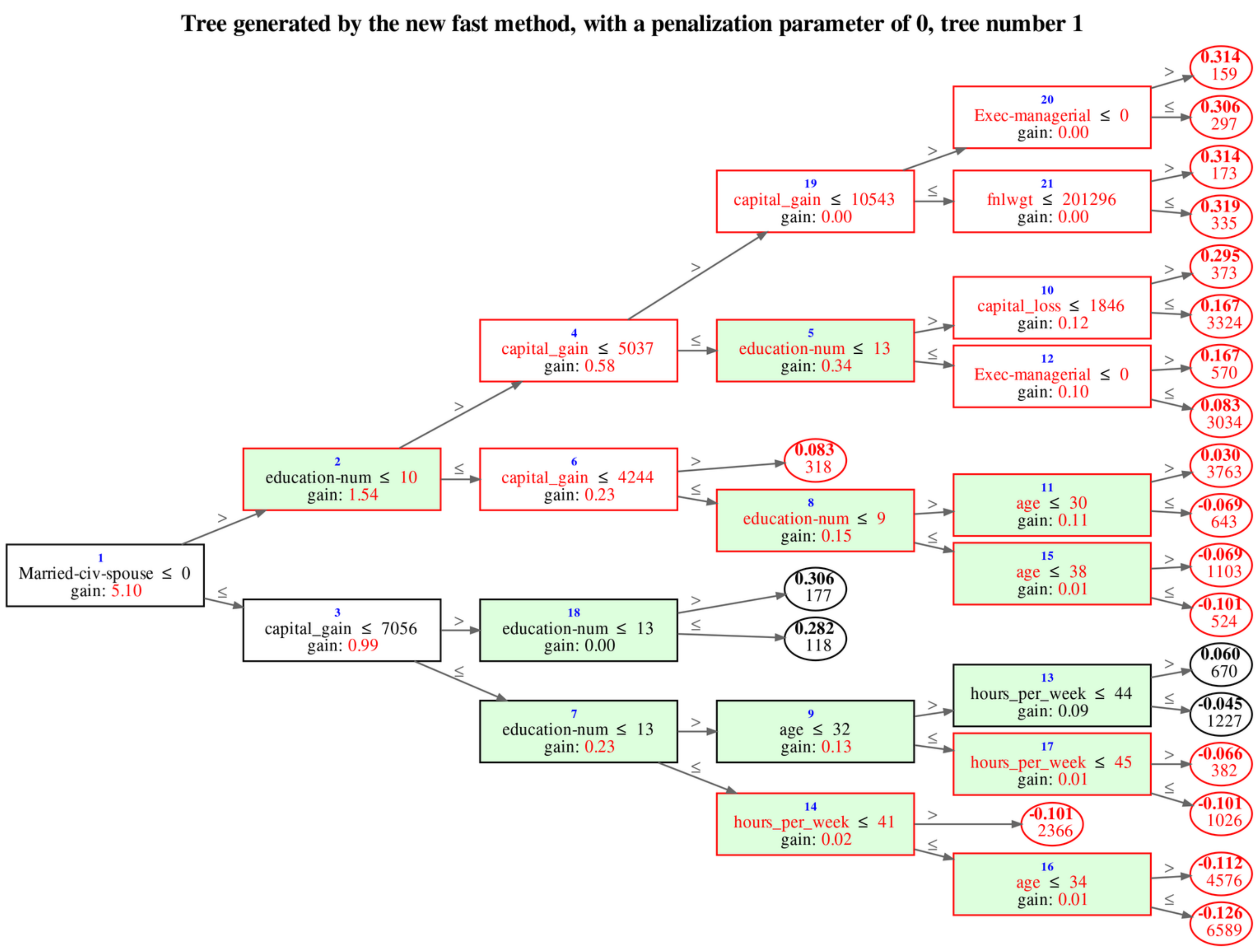}\hfill
\captionsetup{justification=centering,margin=2cm}
\caption{First two trees generated by our new fast method, with no penalization}
\label{tree_fast}
\end{figure}

\begin{figure}[htp]
\centering
\includegraphics[width=1.\textwidth]{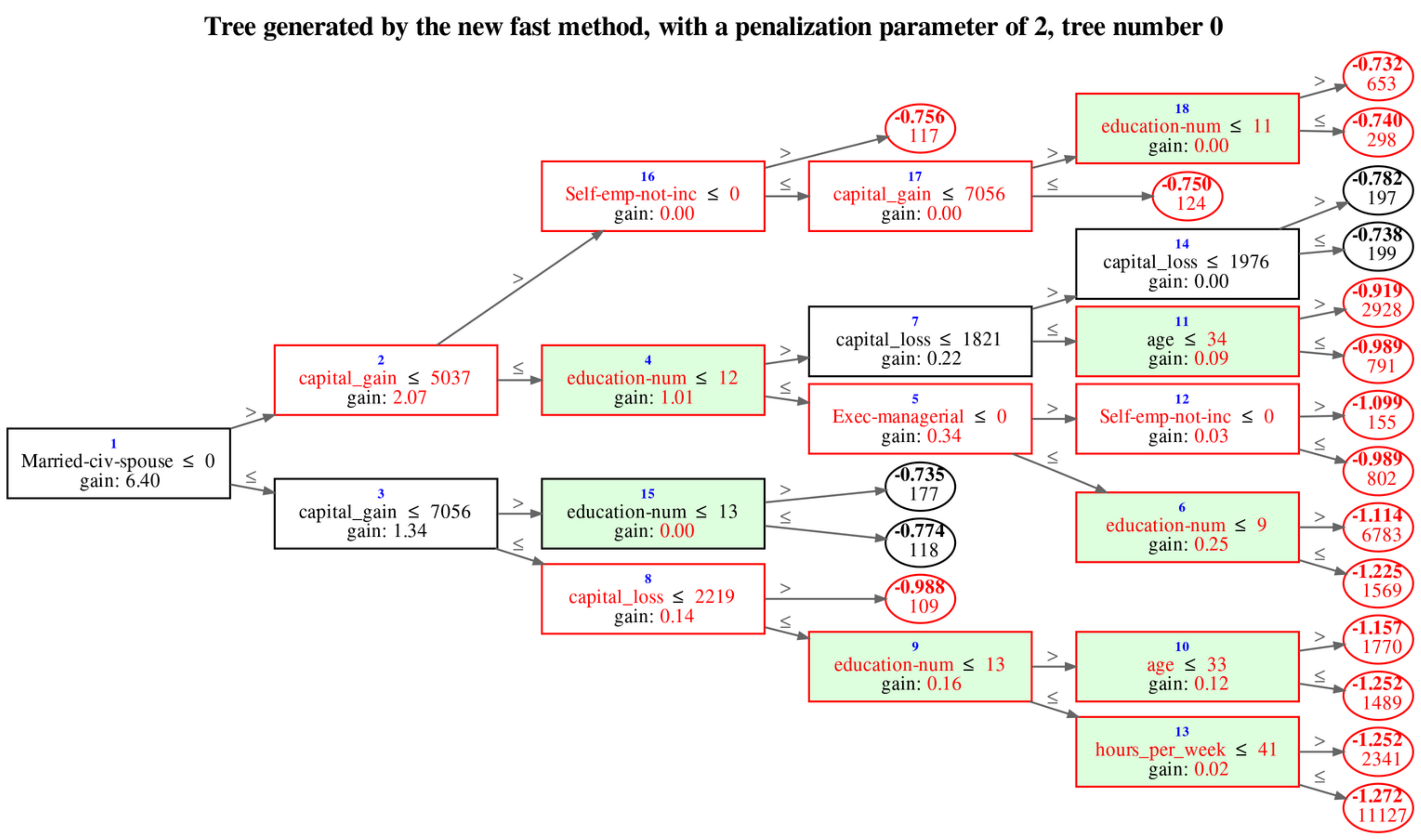}\hfill
\includegraphics[width=1.\textwidth]{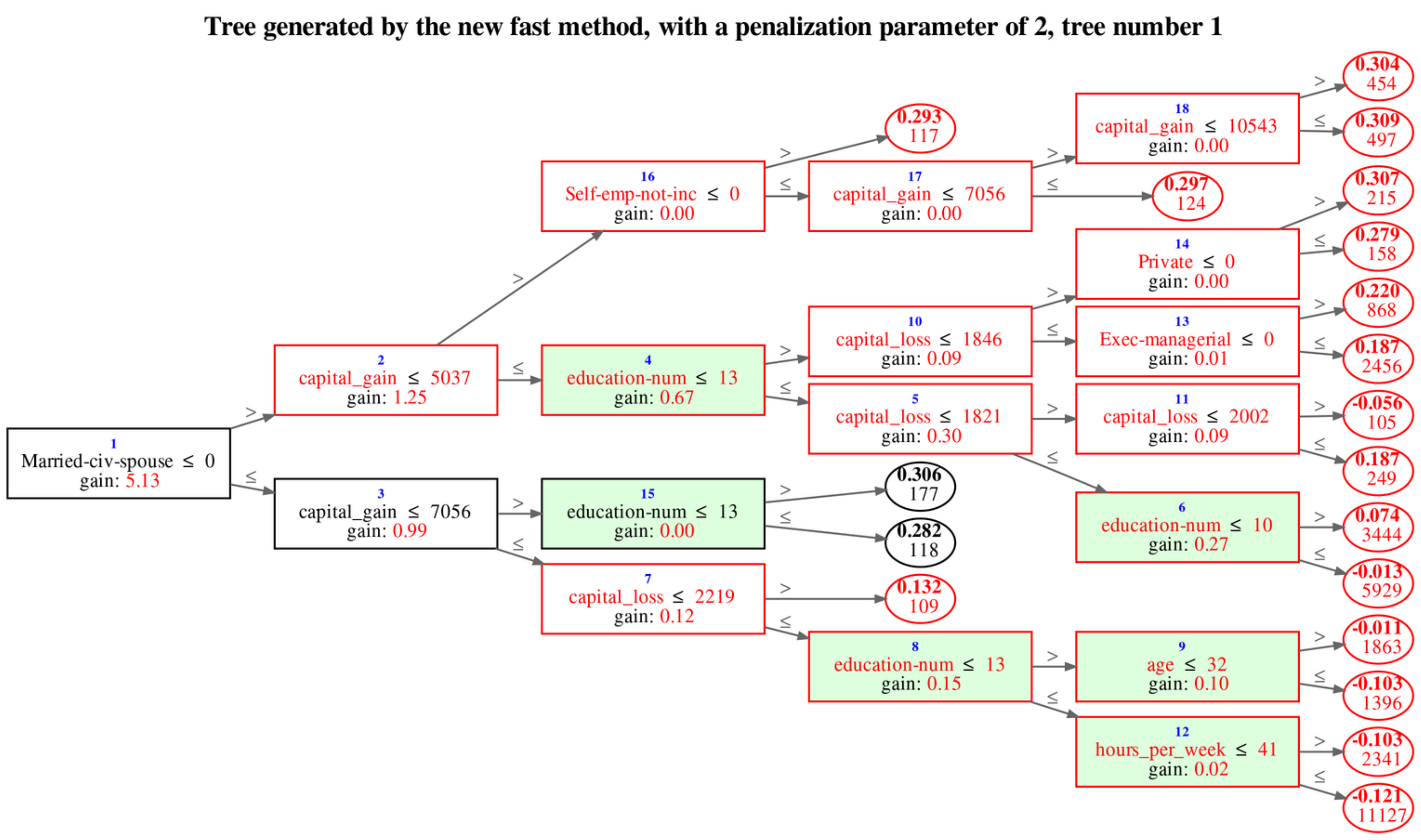}\hfill
\captionsetup{justification=centering,margin=2cm}
\caption{First two trees generated by our new fast method, with a penalization parameter of 2}
\label{tree_fast_penalize}
\end{figure}

On figures \ref{tree_unconstrained}, \ref{tree_lightgbm}, \ref{tree_slow}, \ref{tree_slow_penalize}, \ref{tree_fast} and \ref{tree_fast_penalize}, we
plotted the first trees generated by our different methods to see how they would differ.
Here are a few things to notice to understand the tree visualizations,
\begin{itemize}
    \item The trees are all compared to the unconstrained tree.
    A red leaf means that the leaf does not appear in the unconstrained tree.
    A red node means that the node does not appear in the unconstrained tree (either the splitting feature is different or the threshold is different, or the parent of the node is different).
    A red gain means that the gain for the same node in the unconstrained tree is different;
    \item A green node means that the node is monotonically increasing;
    \item Blue numbers on the nodes show the order in which the nodes have been split.
\end{itemize}

Here are a few observations coming from the comparison of the trees,
\begin{itemize}
    \item As anticipated, the monotone features are extremely important, because they appear in many nodes in the first trees;
    \item Up to a certain point (until the first monotone split), all trees are similar to the unconstrained tree;
    \item For the penalized method, we can indeed see that no monotone split happened in the first levels of the tree;
    \item The outputs of the first tree are all negative because the dataset is imbalanced and there are a lot more 0's than 1's;
    \item The trees get even more different as the algorithms keep running;
    \item Gains for the unconstrained method seem generally greater than for the rest of the methods, which makes sense.
    Also, the unconstrained method has more leaves on this example, which is also coherent.
\end{itemize}

    \section{Explanations of the new methods}
    \subsection{New ways of constraining trees}

We developed 2 new ways of constraining trees.
Both yield better results than the current LightGBM constraining method.
One of the method is about as fast as the current LightGBM.
The other method is much slower (can be as much as twice as slow in extreme worst case scenarii), but is also more accurate.
In this section, we will present both. \\

Both methods originally stem from the observation that the current LightGBM constraining method is over-constraining the trees.
In our methods, when we make a monotone split, instead of imposing constraints on both children that make the outputs of all their descendants
mutually exclusive, as LightGBM currently does by setting a mid point,
we chose to set up only the necessary constraints every time so that the next split would not break the monotone rules.
More precisely, this means that if the split is monotonically increasing, then the left child gets the right output as a maximum constraint,
and the right child gets the left output as a minimum constraint.
However, is the split is monotonically decreasing, then it means that the left child gets the right output as a minimum
constraint and the right child gets the left output as a maximum constraint constraint. \\

Then, when we make any split (monotone or not) in a branch having a monotone node as a parent somewhere, after making the split,
we need to check that the new outputs are not violating any constraint on other leaves of the tree.
The general idea is that we should start from the node where a split was just made, go up the tree,
and every time a monotone node is encountered, we should go down in the opposite branch and check that
the constraints and the new outputs from the new split are compatible.
If they are not, then the constraints need to be updated.
Therefore making a split in a branch can very well update the constraints of other leaves in another branch.

    \subsubsection{The computationally efficient way of constraining trees}

In this method, only one minimum constraint and one maximum constraint need to be stored for each leaf.
A constraint is computed as the extremum of all constraints applied sequentially on a leaf during the construction of the tree.
That makes the constraints very easy to compute, because when a constraint of a leaf has to be updated, it can be done only by performing a minimum or a maximum.
This method is better than the current LightGBM because it doesn't create "gaps" between leaves, and it is still efficient,
because the only additional thing we need is going through the tree and recomputing some split sometimes, but usually not that much
(especially when we compare it to the time spent building histograms).

    \subsubsection{The slower yet more accurate way of constraining the trees}

This method is based on the same principle except that for each leaf, for each feature, for each threshold, we are
going to store one minimum and one maximum constraints.
By doing that, we are going to have many different constraints for every leaf (at most number of features times number of thresholds
times 2 for minimum and maximum constraints per leaf, but usually, it is quite a lot less than that).
Then when a split has to be made with respect to a feature, we can have different left and right constraints for the
children depending on where the constraints were applied on the original leaf.
Additionally, when a split is made, and we start updating the constraints of other leaves, we take into account the fact
that the split may very well have "unconstrained" some leaves (as well as constrained others).
By doing that we are sometimes able to split leaves that would not be splittable using the previous methods.
Moreover, since we compute the exact constraints needed everywhere at every iteration of building the trees,
we are not over-constraining any leaf, and therefore trees can get much more accurate.
However, when using this method, we need to store lots of constraints that are going to be updated very often.
Furthermore, when a leaf is "unconstrained", then all the constraints need to be computed from the beginning
(otherwise we would need to store which leaves the constraints come from, and keep them sorted, which, in our opinion, would not be better).
Because of that, this method is quite a lot slower, even though we tried to implement many optimizations so
that it would remain tractable.

\subsection{Theoretical example}
\begin{figure}
\pgfmathsetseed{1}\begin{subfigure}[b]{\linewidth}
  \centering
  \begin{tikzpicture}
    \draw (0,0) -- (4,0) -- (4,4) -- (0,4) -- cycle;
    \draw (2, 0) -- (2,4);
    \node[below] at (5,4) {True values};
    \draw[fill=blue!35] (5,1.4) circle (0.1cm) node {};
    \node[right] at (5,1.4) {0.7};
    \draw[fill=green!35] (5,2.0) circle (0.1cm) node {};
    \node[right] at (5,2.0) {0.5};
    \draw[fill=red!35] (5,2.6) circle (0.1cm) node {};
    \node[right] at (5,2.6) {0.8};
    \draw[fill=black!35] (5,3.2) circle (0.1cm) node {};
    \node[right] at (5,3.2) {0.2};
    \draw[->] (0,4.3) -- (4,4.3) node[above] {$\text{monotone increasing}$};
    \draw[->] (-0.3, 0) -- (-0.3, 4) node[left] {$\text{unconstrained}$};

    \foreach \x in {1,...,10}
    {
      \pgfmathrandominteger{\a}{210}{390}
      \pgfmathrandominteger{\b}{10}{190}
      \fill[fill=red!35] (\a*0.01,\b*0.01) circle (0.1);
    };
          \foreach \x in {1,...,10}
    {
      \pgfmathrandominteger{\a}{210}{390}
      \pgfmathrandominteger{\b}{210}{390}
      \fill[fill=green!35] (\a*0.01,\b*0.01) circle (0.1);
    };
          \foreach \x in {1,...,10}
    {
      \pgfmathrandominteger{\a}{10}{190}
      \pgfmathrandominteger{\b}{10}{130}
      \fill[fill=blue!35] (\a*0.01,\b*0.01) circle (0.1);
    };
          \foreach \x in {1,...,10}
    {
      \pgfmathrandominteger{\a}{10}{190}
      \pgfmathrandominteger{\b}{210}{390}
      \fill[fill=black!35] (\a*0.01,\b*0.01) circle (0.1);
    };
    \node at (1,2) {\large\textbf{0.3}};
    \node at (3,2) {0.6};
  \end{tikzpicture}
    \caption{First split}\label{a}
  \end{subfigure}

\begin{subfigure}[b]{0.5\linewidth}
      \centering
      \pgfmathsetseed{1}
  \begin{tikzpicture}
    \draw (0,0) -- (4,0) -- (4,4) -- (0,4) -- cycle;
    \draw (2, 0) -- (2,4);
    \draw (2, 2) -- (4,2);

    \foreach \x in {1,...,10}
    {
      \pgfmathrandominteger{\a}{210}{390}
      \pgfmathrandominteger{\b}{10}{190}
      \fill[fill=red!35] (\a*0.01,\b*0.01) circle (0.1);
    };
          \foreach \x in {1,...,10}
    {
      \pgfmathrandominteger{\a}{210}{390}
      \pgfmathrandominteger{\b}{210}{390}
      \fill[fill=green!35] (\a*0.01,\b*0.01) circle (0.1);
    };
          \foreach \x in {1,...,10}
    {
      \pgfmathrandominteger{\a}{10}{190}
      \pgfmathrandominteger{\b}{10}{130}
      \fill[fill=blue!35] (\a*0.01,\b*0.01) circle (0.1);
    };
          \foreach \x in {1,...,10}
    {
      \pgfmathrandominteger{\a}{10}{190}
      \pgfmathrandominteger{\b}{210}{390}
      \fill[fill=black!35] (\a*0.01,\b*0.01) circle (0.1);
    };
    \node at (1,2) {\large\textbf{0.3}};
    \node at (3,3) {0.5};
    \node at (3,1) {0.8};
  \end{tikzpicture}
    \caption{Second split}\label{b}
  \end{subfigure}
\begin{subfigure}[b]{0.5\linewidth}
      \centering
    \pgfmathsetseed{1}
  \begin{tikzpicture}
    \draw (0,0) -- (4,0) -- (4,4) -- (0,4) -- cycle;
    \draw (2, 0) -- (2,4);
    \draw (2, 2) -- (4,2);
    \draw (0, 1.5) -- (2,1.5);

    \foreach \x in {1,...,10}
    {
      \pgfmathrandominteger{\a}{210}{390}
      \pgfmathrandominteger{\b}{10}{190}
      \fill[fill=red!35] (\a*0.01,\b*0.01) circle (0.1);
    };
          \foreach \x in {1,...,10}
    {
      \pgfmathrandominteger{\a}{210}{390}
      \pgfmathrandominteger{\b}{210}{390}
      \fill[fill=green!35] (\a*0.01,\b*0.01) circle (0.1);
    };
          \foreach \x in {1,...,10}
    {
      \pgfmathrandominteger{\a}{10}{190}
      \pgfmathrandominteger{\b}{10}{130}
      \fill[fill=blue!35] (\a*0.01,\b*0.01) circle (0.1);
    };
          \foreach \x in {1,...,10}
    {
      \pgfmathrandominteger{\a}{10}{190}
      \pgfmathrandominteger{\b}{210}{390}
      \fill[fill=black!35] (\a*0.01,\b*0.01) circle (0.1);
    };
    \node at (1,3) {0.2};
    \node at (1,1) {\large\textbf{0.45}};
    \node at (3,3) {0.5};
    \node at (3,1) {0.8};
  \end{tikzpicture}
    \caption{Third split LightGBM}\label{c}
  \end{subfigure}
    \pgfmathsetseed{1}
\begin{subfigure}[b]{0.5\linewidth}
      \centering
  \begin{tikzpicture}
    \draw (0,0) -- (4,0) -- (4,4) -- (0,4) -- cycle;
    \draw (2, 0) -- (2,4);
    \draw (2, 2) -- (4,2);
    \draw (0, 1.5) -- (2,1.5);

    \foreach \x in {1,...,10}
    {
      \pgfmathrandominteger{\a}{210}{390}
      \pgfmathrandominteger{\b}{10}{190}
      \fill[fill=red!35] (\a*0.01,\b*0.01) circle (0.1);
    };
          \foreach \x in {1,...,10}
    {
      \pgfmathrandominteger{\a}{210}{390}
      \pgfmathrandominteger{\b}{210}{390}
      \fill[fill=green!35] (\a*0.01,\b*0.01) circle (0.1);
    };
          \foreach \x in {1,...,10}
    {
      \pgfmathrandominteger{\a}{10}{190}
      \pgfmathrandominteger{\b}{10}{130}
      \fill[fill=blue!35] (\a*0.01,\b*0.01) circle (0.1);
    };
          \foreach \x in {1,...,10}
    {
      \pgfmathrandominteger{\a}{10}{190}
      \pgfmathrandominteger{\b}{210}{390}
      \fill[fill=black!35] (\a*0.01,\b*0.01) circle (0.1);
    };
   \node at (1,3) {0.2};
    \node at (1,1) {\large\textbf{0.5}};
    \node at (3,3) {0.5};
    \node at (3,1) {0.8};
  \end{tikzpicture}
    \caption{Third split fast method}\label{d}
  \end{subfigure}
\begin{subfigure}[b]{0.5\linewidth}
      \centering
    \pgfmathsetseed{1}
  \begin{tikzpicture}
    \draw (0,0) -- (4,0) -- (4,4) -- (0,4) -- cycle;
    \draw (2, 0) -- (2,4);
    \draw (2, 2) -- (4,2);
    \draw (0, 1.5) -- (2,1.5);

    \foreach \x in {1,...,10}
    {
      \pgfmathrandominteger{\a}{210}{390}
      \pgfmathrandominteger{\b}{10}{190}
      \fill[fill=red!35] (\a*0.01,\b*0.01) circle (0.1);
    };
          \foreach \x in {1,...,10}
    {
      \pgfmathrandominteger{\a}{210}{390}
      \pgfmathrandominteger{\b}{210}{390}
      \fill[fill=green!35] (\a*0.01,\b*0.01) circle (0.1);
    };
          \foreach \x in {1,...,10}
    {
      \pgfmathrandominteger{\a}{10}{190}
      \pgfmathrandominteger{\b}{10}{130}
      \fill[fill=blue!35] (\a*0.01,\b*0.01) circle (0.1);
    };
          \foreach \x in {1,...,10}
    {
      \pgfmathrandominteger{\a}{10}{190}
      \pgfmathrandominteger{\b}{210}{390}
      \fill[fill=black!35] (\a*0.01,\b*0.01) circle (0.1);
    };
   \node at (1,1) {\large\textbf{0.7}};
    \node at (1,3) {0.2};
    \node at (3,3) {0.5};
    \node at (3,1) {0.8};
  \end{tikzpicture}
    \caption{Third split slow method}\label{e}
  \end{subfigure}
\captionsetup{justification=centering}
   \caption{Theoretical example of the behavior of the different constraining methods}
\label{the}
\end{figure}
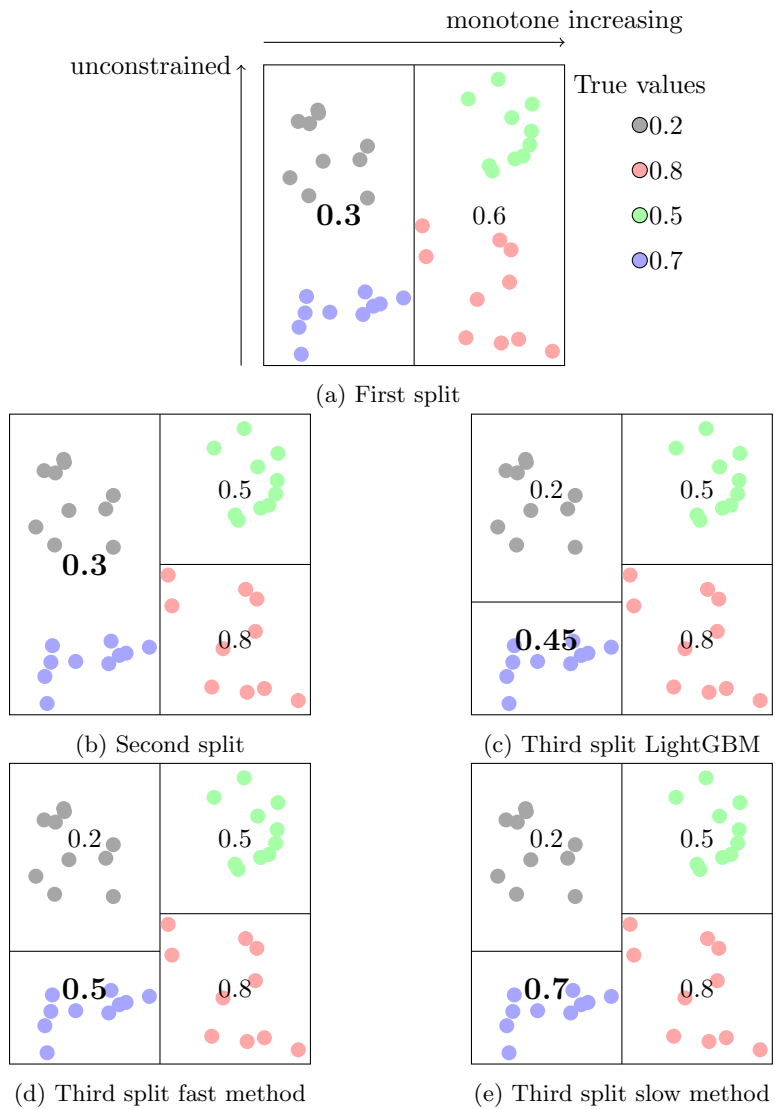

On figure \ref{the}, we imagined a simple situation demonstrating the improvements of our methods.
Let's imagine that we have a regression problem, with the data being represented on figure \ref{a}.
The only constraint is that the output has to be monotonically increasing horizontally.
The true labels of the points are given on the right of figure \ref{a}.
On figure \ref{a}, we take it as given that the first split is the same with all methods and that it yields the values written on the graph.
On figure \ref{b}, we assume that the second split is again the same with all methods and that it yields the values written on the graph.
The 3 following figures depict what would happen for the following split with each method.\\
\begin{itemize}
\item On figure \ref{c}, with LightGBM, since the values on the graph from figure \ref{a} were 0.3 and 0.6, the mid point is 0.45.
Therefore, the blue dots are upper bounded and labelled as 0.45.
\item On figure \ref{d}, with the fast method, once the first split is made, the leaf on the left is given an upper bound of 0.6.
Once the second split is made, every leaf on the left is given an upper bound of 0.5.
Therefore, the blue dots are labelled as 0.5
\item On figure \ref{e}, with the slow method, once the first split is made, the leaf on the left is given an upper bound of 0.6.
Once the second split is made, the top part of the left leaf is given an upper bound of 0.5, and the lower part of the left leaf is given an upper bound of 0.8.
Therefore, the blue dots are labelled as 0.7, their true value.
\end{itemize}

We can therefore see that our methods are better at this specific regression problem, the slow method achieving optimal results with only one tree.
Similar examples happen very often both in regression and classification tasks, when building trees.
Therefore, our methods should consistently yield better results than the current LightGBM implementation.

    \subsection{A heuristic penalty for monotone splits}

\subsubsection{Why penalizing monotone splits is a good idea}

The main drawback of adding monotone constraints to a tree is that it may end up
over-constraining the tree because of the greediness of the tree splitting algorithm, and therefore loose in predictive power.
By building a tree in a greedy fashion, we don't know how much the constraints of monotone splits are going to affect the following nodes
(it may reduce the gain of the children by a lot).
For our new methods, we don't know either by how by much a new split is going to
constrain other surrounding leaves and their future children (in the event that the leaf
we are splitting is part of a branch growing from a monotone node). \\

When using our new method, even though it already yields good results as we will see in section \ref{res},
both these effect happen in almost any tree with reasonable depth when making at least one monotone split.
However, we do not quantify or take into account any of these 2 drawbacks at all in the algorithm.
Yet they can be huge at the first levels of the trees, because the constraints are going to affect most of the leaves.
Therefore, amongst other things, we are very likely to overestimate the gain generated by splitting a leaf on a monotone feature;
especially early in the trees, where the constraints are going to impact many leaves. \\

Unfortunately, this reduction of gain can't be computed greedily.
Indeed, we would need to know what is going to happen after to quantify it (or we could try to predict it but that is not an easy task either).
However, we were able to develop a simple heuristic that improves the results, by penalizing monotone splits early when building trees. \\

The general idea of the heuristic is that we are likely to not want to make a monotone split in the first levels of a tree
because this is going to constrain too many leaves.
Then, generally, the deeper we go, the more inclined to monotone spits we are.
This is a very general rule than can have many exceptions, but we believe it is true more often than not.
Therefore, we designed a penalty function that is a function of depth, that will multiply the gain of monotone splits at a given depth.
It is parametrised by only one parameter
that we will call $\gamma$ here for simplicity, and that allows monotone splits to be penalized the higher they appear in a tree. \\

\subsubsection{Penalization formula and usage}

The final penalty $p$ that will multiply he gain of a node depends on the tunable parameter $\gamma$ and on the depth of the node in the tree $d$, depth 0 being the root.
So we have $0 \leq d \leq \text{max depth}$.
The penalty is computed according to the following formula,
$$p = \left\{
    \begin{array}{ll}
        0 & \mbox{if } \gamma \geq d + 1 \\
        1 - \frac{\gamma}{2^{d}} & \mbox{if } \gamma \leq 1 \\
        1 - 2^{\gamma - 1 - d} & \mbox{else}
    \end{array}
\right.$$
Moreover, when using the LightGBM framework, we add an extremely small $\epsilon$ to $p$, because only splits with a strictly positive gain can be performed,
and in the event that we would only have monotone splits available, we still want split one of them. \\

On figure \ref{penalization_function} we plotted what the penalty looks like for different penalization parameters, as a function of depth.
There are some important things to mention about this penalty function:
\begin{itemize}
    \item When $\gamma=0$, then there is no penalty at all;
    \item When $\gamma \in [0; 1]$, then we gradually penalize all depths, and the closer to 1 they are, the more they are penalized;
    \item When $\gamma \in [1; 2]$, then the penalty for the first level is 0 (we prohibit any monotone split on this level),
    and we keep gradually penalizing monotone splits at all depths, and the closer to depth 2 they are, the more they are penalized;
    \item When $\gamma \in [2; 3]$, then the penalty for the first two levels is 0 (we prohibit any monotone split on these levels),
    and we keep gradually penalizing monotone splits at all depths, and the closer to depth 3 they are, the more they are penalized;
    \item ...
\end{itemize}
Therefore, the parameter $\gamma$ is a very intuitive parameter that shouldn't confuse the end user.

\subsubsection{Penalization vs. depth plot}
\begin{figure}[htp]
\centering
\includegraphics[width=1.\textwidth]{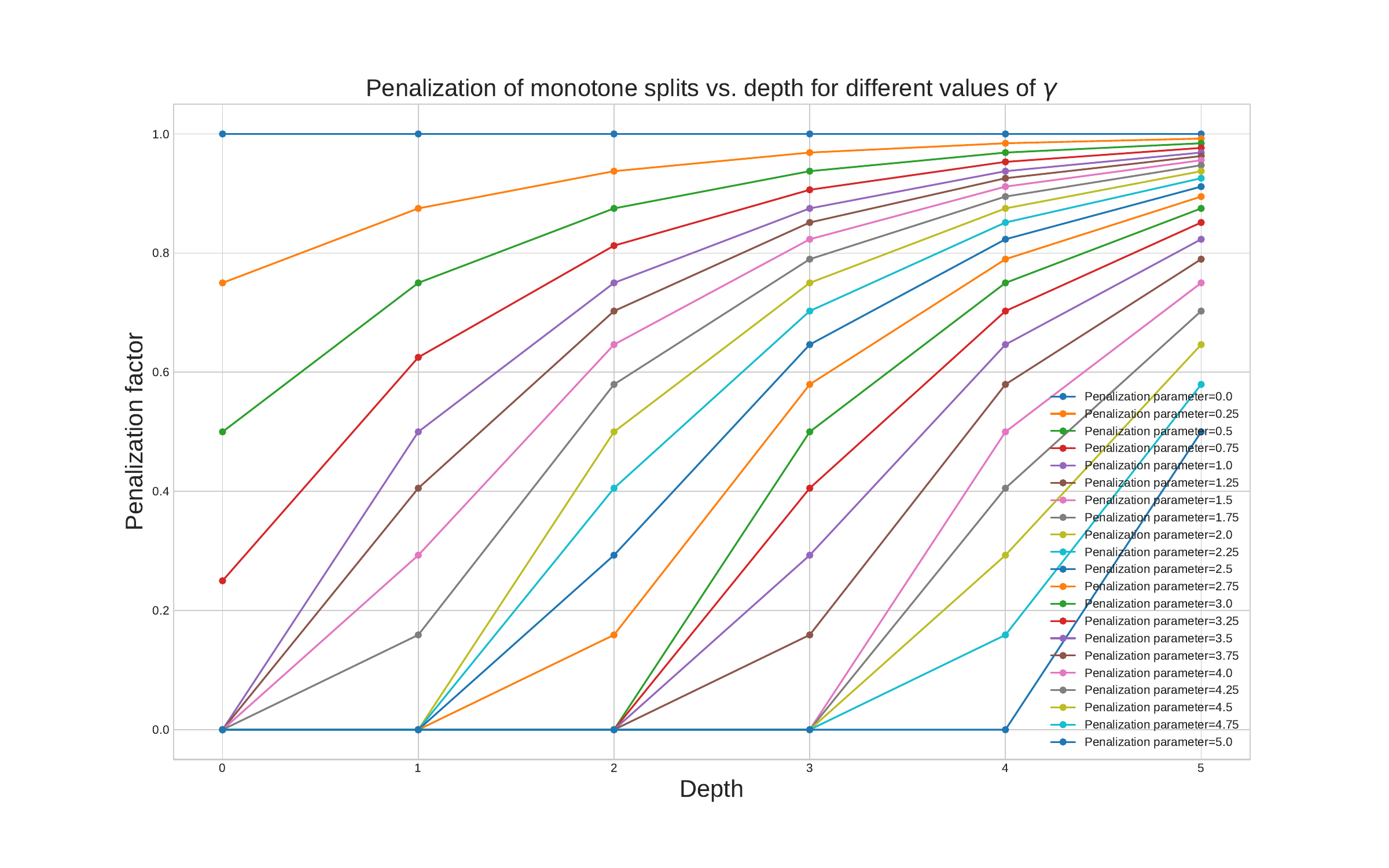}\hfill
\captionsetup{justification=centering,margin=2cm}
\caption{Penalty function for different penalization parameters}
\label{penalization_function}
\end{figure}

On figure \ref{penalization_function}, we plot the penalty factor that will multiply the gain of monotone splits,
for different depths and penalization parameters.

\newpage

    \section{Conclusion}
    In this report, we propose two new methods to enforce monotonic constraints in regression and classification trees, as well as a heuristic to improve the results.
    During our tests, we found that our new methods consistently achieve better results than the current LightGBM constraining method.
    Our fastest method does not generate significant additional computational cost compared to the current LightGBM.
    Therefore, we believe that our work should replace the current LightGBM implementation.
    Finally, as mentioned previously, our work is not LightGBM-specific and could be implemented in any regression or
    classification tree.

\end{document}